\documentclass{article}

\usepackage{arxiv}

\usepackage[utf8]{inputenc} % allow utf-8 input
\usepackage[T1]{fontenc}    % use 8-bit T1 fonts
\usepackage{hyperref}       % hyperlinks
\usepackage{url}            % simple URL typesetting
\usepackage{booktabs}       % professional-quality tables
\usepackage{amsmath}        % mathematical operators and environments
\usepackage{amsfonts}       % blackboard math symbols
\usepackage{amssymb}
\usepackage{algorithm}
\usepackage{algorithmic}
\usepackage{nicefrac}       % compact symbols for 1/2, etc.
\usepackage{microtype}      % microtypography
\usepackage{lipsum}    % Can be removed after putting your text content
\usepackage{graphicx}
\usepackage[numbers,sort&compress]{natbib}
\usepackage{doi}

\title{SpectRAG: Weighted Chunk Graphs for Token-Budgeted Context Selection in Multi-Hop Retrieval-Augmented Generation}

\author{
  \href{https://orcid.org/0000-0002-7655-1840}{\includegraphics[scale=0.06]{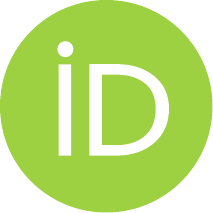}\hspace{1mm}Zafer Sahinoglu} \\
  Mitsubishi Electric US\\
  Cambridge, 5900-A Katella Avenue \\
  California 90630, USA \\
  \texttt{zafer.sahinoglu@meus.com} \\
  \And
  \href{https://orcid.org/0009-0009-3030-6530}{\includegraphics[scale=0.06]{orcid.pdf}\hspace{1mm}Abdulazeez Mohammed} \\
  Mitsubishi Electric Innovation Center\\
  (MELIC) \\
  Cambridge, MA 02139, USA\\
  \texttt{abdulazeez\_mohammed@outlook.com} \\
  \And
  \href{https://orcid.org/0000-0002-8690-2042}{\includegraphics[scale=0.06]{orcid.pdf}\hspace{1mm}Buse Yaren Kazangirler} \\
  Mitsubishi Electric Innovation Center\\
  (MELIC) \\
  Cambridge, MA 02139, USA\\
  \texttt{buseyarentekin@outlook.com} \\
  \And
  \href{https://orcid.org/0000-0002-2670-0940}{\includegraphics[scale=0.06]{orcid.pdf}\hspace{1mm}Cahit Berkay Kazangirler} \\
  Institute of Graduate Programs,\\
  Dept. of Computer Engineering,\\
  Karabuk University, Karabuk, 78050, Turkey\\
  \texttt{cahitberkaykazangirler@outlook.com} \\
}

\renewcommand{\shorttitle}{\textit{arXiv} Template}

\hypersetup{
  pdftitle={A template for the arxiv style},
  pdfsubject={q-bio.NC, q-bio.QM},
  pdfauthor={David S.~Hippocampus, Elias D.~Striatum},
  pdfkeywords={First keyword, Second keyword, More},
  urlcolor=blue,
}

\begin{document}
\maketitle

\begin{abstract}
  Retrieval-augmented generation (RAG) systems ground large language models (LLMs) by retrieving external evidence and conditioning generation on the retrieved context. Although dense retrieval is effective for local semantic matching, independently ranked retrieval can return redundant passages, miss cross-chunk dependencies, and provide incomplete evidence for multi-hop reasoning. We propose \textit{SpectRAG}, a spectral graph-augmented retrieval framework that combines hybrid candidate seeding, weighted chunk graph construction, graph-based expansion, spectral relevance propagation, and token-budgeted context selection. Document chunks are represented by dense semantic embeddings and graph-spectral coordinates derived from a weighted chunk graph whose edges encode semantic similarity, document adjacency, section hierarchy, shared entities, citation links, and similarity between LLM-generated utility questions. At query time, SpectRAG builds a high-recall seed set using exact inner-product search and hybrid relevance signals, expands it over the chunk graph, re-scores candidates using graph proximity and spectral information, and assembles the final context under a fixed token budget. The goal is not merely to retrieve individually similar chunks, but to construct a context that is relevant, diverse, structurally connected, and sufficient for grounded reasoning. Experiments on MultiHopRAG, UltraDomain, and long-document question-answering settings show that SpectRAG improves broad evidence recovery and achieves competitive generation quality against dense, hybrid, HyDE, LightRAG, and NaiveRAG baselines. The results further indicate that graph-spectral structure is particularly useful for inference-oriented and cross-document retrieval, while highlighting trade-offs in early precision, computational cost, and domain-dependent performance.
\end{abstract}

\keywords{Retrieval-augmented generation \and token-budget context selection \and graph-based retrieval \and spectral graph theory \and dense vector search \and multi-hop reasoning \and passage diversity}

\section{Introduction}
\label{sec:introduction}
Retrieval-augmented generation (RAG) has become a practical architecture for grounding large language models (LLMs) in external corpora \cite{b1}. In a standard pipeline, documents are chunked, chunks are embedded, a query is embedded, and the top-$k$ chunks most similar to the query are provided as context to an LLM. Dense passage retrieval \cite{b2} and hybrid sparse--dense retrieval are now widely used because they are scalable and effective for local factual lookup. In this paper, we see this constraint as a separation of two paradigms for retrieval. One of them is the classical top-$k$ paradigm, where chunks are scored based on their individual similarities to the query, and the top-scoring chunks are selected. The other paradigm is called context-set retrieval, where the goal is to find a set of chunks that, together, provide enough evidence for generation. In this paradigm, a good RAG retriever should not only be able to retrieve chunks similar to the query, but also form a cohesive context set. Nevertheless, the method of extracting the top-$k$ vectors is subject to three recurring limitations. Firstly, given that adjacent or semantically similar passages generally occupy the same region of the embedding space, this method has the potential to eliminate superfluous evidence. Secondly, it may overlook structurally important segments that are necessary for multi-step reasoning, even if they are not immediately close to the query. Thirdly, while this method treats segments as largely independent units, real-world text corpora possess a structure that includes document order, section hierarchy, shared entities, citations, temporal dependencies, and conceptual links \cite{b3}.

While the standard top-k retrieval method is practically effective, it is limited when the answer requires a context set rather than a group of similar passages. For dense vector retrieval, chunks are usually ranked based on their individual similarity to the query. The success of this ranking mechanism depends on the local representation of the required evidence in only one or a few semantically similar passages. However, many real-world question-answering (QA) and reasoning tasks require evidence distributed across multiple chunks, documents, or sections. In such cases, the most advantageous setting of context may not be composed of the highest-ranked components alone, but of passages that are jointly relevant, complementary, and structurally interconnected \cite{b4}--\cite{b6}.

Graph-based retrieval methods have emerged as a response to these limitations. Retrieval-augmented generation with graphs (GraphRAG)-style approaches represent information as graphs and exploit relational links to support retrieval and generation across connected evidence \cite{b6}. Related work has explored entity graphs, document graphs, knowledge graphs, and graph-augmented memory structures. For example, GEM-RAG \cite{rappazzo2024gemrag} graphical eigenmemories by tagging data segments with utility questions, connecting them into a graph, and using eigendecomposition to form higher-level memory nodes. CrossAug further highlights that many GraphRAG pipelines miss relations whose supporting evidence spans multiple chunks, and proposes an offline graph-augmentation strategy to recover cross-chunk structure \cite{b7}. These studies indicate that retrieval can benefit from modeling both the semantic similarity between a query and individual chunks and the structural relationships among these chunks.

However, existing graph-enhanced RAG methods often focus on graph construction, entity-level relations, or community-level summarization. The problem of selecting a compact, diverse, and structurally connected context set remains underexplored. A retrieval system for multi-hop and long-document reasoning should, in principle, be designed to satisfy numerous objectives simultaneously. The goals of this initiative are numerous and include, but are not limited to:
\begin{itemize}
  \item To reduce redundancy, resulting in more efficient and varied retrieval strategies and, hence, greater confidence in the ongoing efforts.
  \item Bridge evidence recovery in process.
  \item Ensure structural coherence of the selected segments is very important.
\end{itemize}

\subsection{Motivation Behind The Study}
We introduce \emph{SpectRAG}, a retrieval framework that combines dense candidate seeding with graph-spectral context-set optimization. This framework does not thoroughly replace dense retrieval techniques. Rather, it uses dense vector similarity to generate an initial candidate pool with high recall, then leverages the structure of a weighted chunk graph to refine the final context set. The suggested framework represents chunks using both dense semantic embeddings and spectral coordinates that are derived from the graph Laplacian. These spectral coordinates offer structural insights into the corpus retriever to consider graph connectivity, latent communities, and inter-chunk relationships during the selection process, and enable the retriever to account for graph connectivity, latent communities, and cross-chunk relations during selection.

In the querying stage, SpectRAG converts the user query into embeddings, and then creates a seed set from the embedding space that has high recall. The seed set is then further extended via graph neighborhoods for fetching other related segments that do not necessarily match the query well, but can act as connectors or be evidence itself. The relevance of queries is then propagated to the candidate subgraph that is affected by this, thus making relevance propagate from the matched segments to other related evidence. Finally, SpectRAG applies a token-budgeted context selection stage that assembles the final context under a fixed input budget. This stage prioritizes relevant and structurally connected chunks, preserves a protected core of high-confidence evidence, applies a precision-oriented filtering strategy for lower-ranked candidates, and backfills the remaining budget when necessary to avoid under-utilization of the available context window.

\subsection{Limitations of Top-k Dense Retrieval}
\label{sec:limitations}
In conventional dense retrieval, each chunk is scored independently according to its similarity to the query. Given a corpus of chunks $V={c_1,c_2,\ldots,c_n}$, chunk embeddings $\mathbf{e}_i \in \mathbb{R}^{d}$, and a query embedding $\mathbf{e}q \in \mathbb{R}^{d}$, the standard dense retrieval baseline can be written as
\begin{equation}
  S_k^{\mathrm{dense}}(q)
  =
  \operatorname{TopK}{c_i \in V}^{k}
  \operatorname{sim}(\mathbf{e}_q,\mathbf{e}_i),
\end{equation}
where $\operatorname{sim}(\cdot,\cdot)$ is typically cosine similarity. This setup is computationally efficient and effective as long as the key evidence lies in a single or a few semantically similar chunks \cite{b1, b2}. This approach assumes that the best context to generate from can be chosen by ranking the chunks themselves. It might be wrong if the response needs complementary, concise evidence.

Redundancy creates a problem because dense embedding vectors are generally grouped into clusters of adjacent or overlapping chunks of text. In other words, in the case of the selection of the top $k$, multiple chunks with identical content will be selected, thereby leaving less space in the context space for different evidence. This becomes a problem in retrieval-augmented generation because of the small size of the context space, leading to fewer pieces of evidence being considered. However, the use of diversity-oriented retrieval can help mitigate the problem \cite{b4}.

The second limitation is the potential loss of bridge evidence.  While answering the question using multi-hop reasoning, the evidence needed for the answer could include some passage chunks bridging entities, events, definitions, references, or claims.  Such bridge chunks are not always directly similar to the user query and may therefore receive lower dense similarity scores. Therefore, independent top-$k$ rankings can retrieve locally relevant passages, but ignore the evidence required to connect them into a total reasoning path. This issue is especially important in benchmarks such as HotpotQA, 2WikiMultiHopQA, and MuSiQue, where supporting evidence is distributed across multiple passages or reasoning steps \cite{b8, b9, b10}.

The third limitation observed in conventional dense search systems is structure blindness. In this type of system, information is divided into separate divisions for retrieval. However, most databases usually have relationships existing between different parts of information. The relationship could cut across many different dimensions, including document order relationships, hierarchical relationships between different sections of the document, relationships between common entities, citation or referencing that ties one part of information to another, time dependencies that show the timing of the information, as well as conceptual relationships.

These limitations suggest that top-$k$ dense retrieval should be treated as a strong baseline rather than as the final retrieval objective. This encourages the proposed SpectRAG framework combines hybrid candidate seeding, graph neighborhood expansion, proximity-based re-scoring, and token-budgeted final context selection.

\subsection{Contributions}
To summarize, the essential contributions of this research are outlined as follows:
\begin{itemize}
  \item SpectRAG, a spectral graph-augmented retrieval framework integrates dense vector search, weighted chunk graph construction, Laplacian spectral coordinates, graph-based candidate expansion, relevance diffusion, and cluster-aware final selection.
  \item A graph-based retrieval pipeline that uses dense retrieval as an initial entry point and then exploits graph structure to recover bridge evidence and reduce redundant context.
  \item An evaluation on MultiHopRAG (retrieval and answer accuracy against gold labels) and four UltraDomain domains (pairwise LLM-judge comparison), with index-cost measurements against LightRAG.
  \item We design an ablation study to isolate the contribution of graph expansion, spectral coordinates, diffusion, graph edge signals, and diversity-aware context selection.
\end{itemize}

\emph{SpectRAG} is a retrieval framework that augments dense vector search with a weighted chunk graph and the spectral structure of its graph Laplacian. Rather than treating dense retrieval and graph-based retrieval as competing alternatives, SpectRAG adopts dense vector search as a high-recall entry point and subsequently exploits graph structure to expand, diffuse, diversify, and select the final context set. Each chunk is defined using both its dense semantic embedding and graph-spectral coordinates derived from a weighted chunk graph. The graph can integrate numerous edge signals, including semantic similarity, document adjacency, section hierarchy, shared entities, citation relations, and similarity between LLM-generated utility queries.

\section{Related Work}
\subsection{Retrieval-Augmented Generation}
RAG is a method which integrates user query with related passages from an external knowledge source \cite{b3,b8}. The system being studied includes two core elements: the \textit{retrieval component} and the \textit{generation component}. The retrieval component identifies relevant passages for the query made by the user, whereas the generation component uses those passages as context to generate a response.

Previous RAG models have demonstrated that the combination of parametric language models with non-parametric external memories works well for knowledge-heavy tasks where the required information cannot be completely represented by the model parameters or where information changes over time. In real-world scenarios, the retrieval process is typically carried out using methods such as sparse retrieval, BM25 retrieval \cite{b11}, dense passage retrieval, or a combination of sparse and dense retrieval.Sparse retrievers perform retrieval based on keyword matching, whereas dense retrievers map the query and documents into a vector space and retrieve semantically similar passages. Although it is effective, traditional RAG generally counts on independently ranked passages. While this design is adequate for local factual lookups, it may prove inadequate when the answer requires evidence distributed across multiple passages, documents, or structural relations.  Recent research has sought to enhance not only the relevance of retrieved passages but also the overall coherence and diversity of the final context, encouraging the audience to feel optimistic about future developments.

\begin{figure*}[t]
  \centering
  \includegraphics[width=0.93\textwidth]{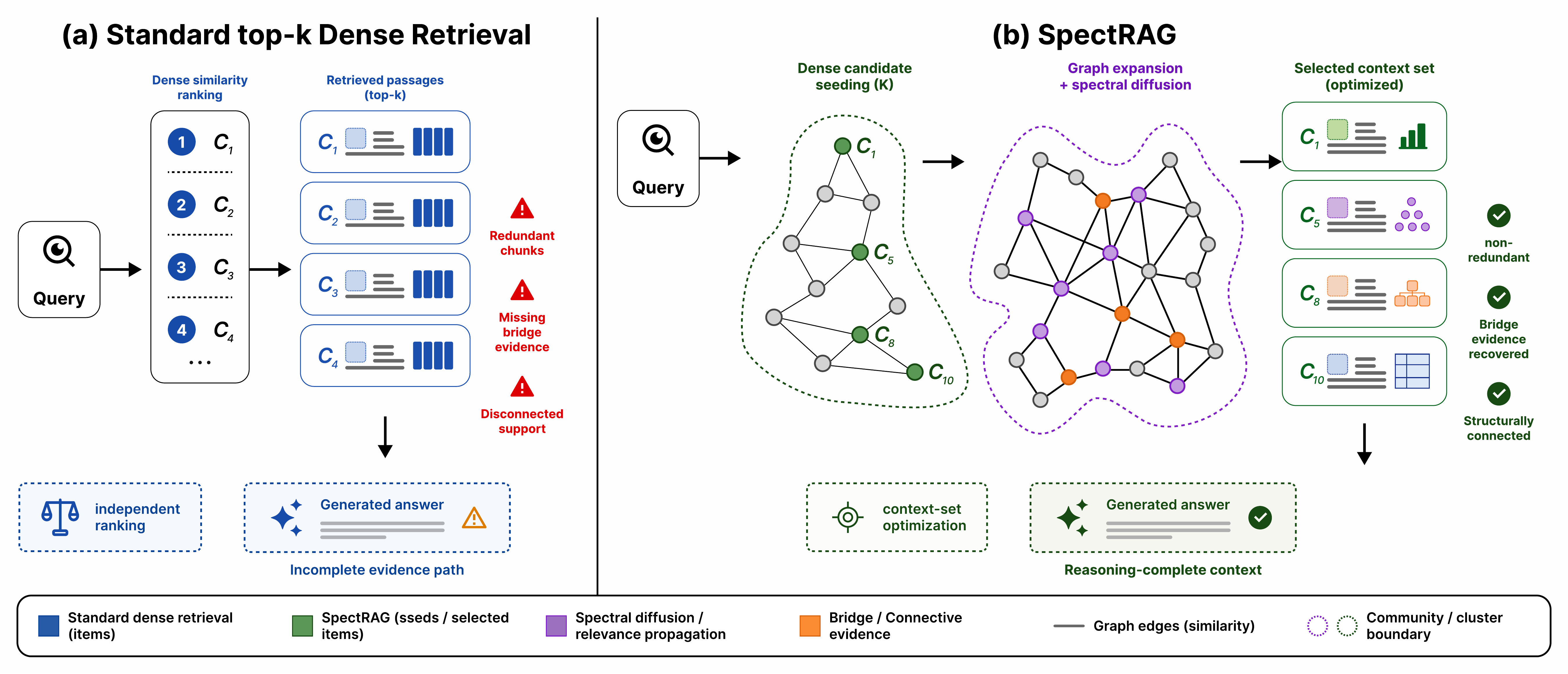}
  \caption{Top-k paradigm vs. our SpectRAG framework}
  \label{fig:dense_vs_spectrag}
\end{figure*}

\subsection{Diversity-Aware and Context-Set Retrieval}
The conventional systems of retrieval concentrate on the relevance of the individual passages but not on the assessment of the relevance of the entire set of retrieved passages. The generator takes the set of passages and the relevance of such a set depends not only on the relevance of the individual passage to the given query but also on the complementarity, non-redundancy, and sufficiency of the passages, making it evident that non-redundancy plays an important role in information retrieval.

In the context of RAG, context-set retrieval is the process of selecting a subset of passages that optimizes set utility. The selected context set needs to preserve query relevance, cover evidence from all sides, avoid repetition, and provide enough data for generation. In multihop reasoning tasks and long-document reasoning, the selected context must also be coherent across related passages, since the answer to the question may depend on connecting evidence from different parts of the corpus \cite{b13}. Relationship-aware table discovery methods, such as SANTOS, similarly show that semantic relationships among structured data elements can improve search beyond column- or item-level matching alone \cite{b14}.

A conceptual comparison between independent dense retrieval and the proposed SpectRAG strategy is shown in Fig.~\ref{fig:dense_vs_spectrag}. In standard dense retrieval, passages are ranked independently based on their similarity to the query, which can lead to redundant passages, missing bridge evidence, and disconnected support. In contrast, SpectRAG constructs the final context by combining relevance, graph connectivity, spectral diffusion and diversity-aware selection under a token-budget constraint.

This perspective is further developed in SpectRAG through token-budgeted context assembly. In contrast to conventional top-ranked retrieval, where the final passages are selected only according to independent similarity scores, SpectRAG constructs the final context under a fixed token budget. Candidate chunks are evaluated not only by relevance, but also by graph proximity, document coverage, redundancy, and their contribution to a compact evidence set. This design allows the system to preserve high-confidence evidence while using the remaining budget for complementary and structurally useful chunks \cite{b15}.

\subsection{Graph-Enhanced Retrieval and GraphRAG}%review--done
GraphRAG methods explicitly represent corpus structure. Edge \emph{et al.} \cite{b12} propose a graph-based approach to query-focused summarization in which an LLM extracts an entity knowledge graph, groups related entities into communities, generates community summaries, and uses these summaries to answer global questions over a corpus. This method addresses questions such as identifying themes or relationships across many documents, where naive local retrieval can fail. Graph-enhanced retrieval is especially relevant when answer evidence is distributed across multiple pieces of text. Zhang \emph{et al.} \cite{b7} argue that existing GraphRAG systems often extract relations within individual chunks, leaving cross-chunk relations systematically absent. Their CrossAug method enriches graph indices with missing cross-chunk relational structure using a topology-aware graph neural network and targeted LLM completion.

\subsection{Weighted Chunk Graph}
The framework constructs a weighted graph $G=(V,E,W)$ whose nodes are chunks. Edge weights combine multiple evidence signals:
\begin{equation}
  w_{ij} =
  \alpha_1 w_{ij}^{sem}
  + \alpha_2 w_{ij}^{adj}
  + \alpha_3 w_{ij}^{entity}
  + \alpha_4 w_{ij}^{utility}
  + \alpha_5 w_{ij}^{citation}.
\end{equation}

\begin{figure*}[h]
  \centering
  \includegraphics[width=0.93\textwidth]{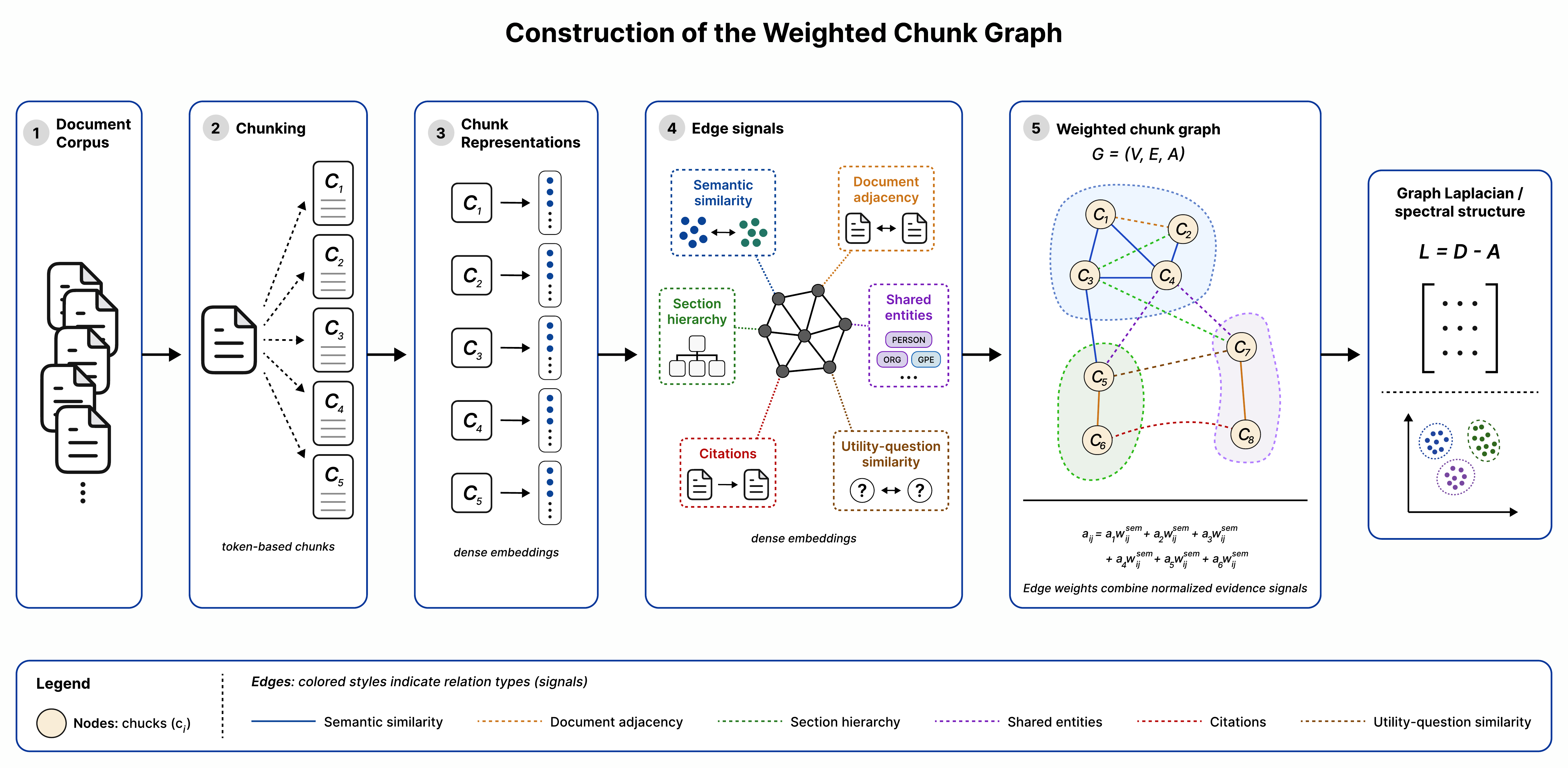}
  \caption{Documents are chunked into token-based passages (1--2) and encoded as dense embeddings (3). Rather than treating chunks as independent vectors, SpectRAG connects them with edges derived from six complementary evidence signals---semantic similarity, document adjacency, section hierarchy, shared entities, citations, and utility-question similarity (4).}
  \label{fig:weighted_chunk_graph}
\end{figure*}

The semantic edge weight is computed from dense embeddings:
\begin{equation}
  w_{ij}^{sem} = \max(0, \cos(e_i,e_j)),
\end{equation}
usually restricted to a sparse $k_g$-nearest-neighbor graph. Document adjacency connects neighboring chunks in the same source document. Section hierarchy connects chunks that share headings or parent sections. Entity edges connect chunks sharing named entities, product names, technical terms, dates, or organizations. Utility-question edges connect chunks that answer similar generated questions, following the intuition that functional similarity may be more retrieval-relevant than surface similarity alone \cite{rappazzo2024gemrag}. Citation edges connect chunks with references, hyperlinks, footnotes, or explicit dependency relations.

Let $A$ denote the weighted adjacency matrix and $D$ the diagonal degree matrix with $D_{ii}=\sum_j A_{ij}$. The unnormalized Laplacian is:
\begin{equation}
  L = D - A,
\end{equation}
and the symmetric normalized Laplacian is:
\begin{equation}
  L_{sym} = I - D^{-1/2} A D^{-1/2}.
\end{equation}

The symmetric normalized Laplacian $L_{\mathrm{sym}}$ of this graph provides the spectral coordinates and diffusion operator used at query time. Edges that connect distinct communities (e.g., $C_5$--$C_6$, $C_7$--$C_8$) carry the bridge evidence that top-$k$ dense retrieval systematically misses.

Fig.~\ref{fig:weighted_chunk_graph} shows the construction of the weighted chunk graph. Nodes are document chunks and edges are multiple relation signals including semantic similarity, document adjacency, section hierarchy, shared entities, citation links and utility-question similarity. These relation signals determine the adjacency matrix for Laplacian computation, spectral coordinate extraction, graph expansion and relevance propagation.

\subsection{Context-Set Retrieval Objective}
\label{sec:objective}

A conventional dense retriever ranks chunks independently by their
similarity to the query. Given a similarity function
$\mathrm{sim}(\cdot,\cdot)$, typically cosine similarity, the initial
relevance score of chunk $c_i$ is
\begin{equation}
  s_0(c_i, q) = \mathrm{sim}(e_q, e_i).
  \label{eq:relevance}
\end{equation}
The standard dense baseline selects the $k$ chunks of highest individual
relevance,
\begin{equation}
  S_k^{\mathrm{dense}}(q)
  = \operatorname*{Top\text{-}K}_{c_i \in V}\, s_0(c_i, q).
  \label{eq:topk}
\end{equation}
Equation~\eqref{eq:topk} scores each chunk in isolation: the marginal value
of a chunk does not depend on what has already been selected. As established in Section~\ref{sec:limitations} and formalized in Section~\ref{sec:objective}, this
independence is the source of the redundancy and missing-bridge-evidence
failures that motivate SpectRAG. We restate it here only to fix the notation
$s_0$ and $S_k^{\mathrm{dense}}$ used in the objective below: whereas
\eqref{eq:topk} optimizes a sum of independent scores, the context-set
objective of~\eqref{eq:context_objective} optimizes over the selected set jointly.

SpectRAG therefore formulates retrieval as a \emph{context-set selection}
problem. The goal is not to select the individually highest-scoring chunks, but to construct a compact context set $C_q \subseteq V$ with $|C_q| = k$ that is
together beneficial for grounded generation:
\begin{equation}
  C_q^{*}
  =
  \arg\max_{C_q \subseteq V,\, |C_q|=k}
  \mathcal{F}(C_q,q),
\end{equation}
where the set-level utility $\mathcal{F}$ is defined as
\begin{equation}
  \begin{split}
    \mathcal{F}(C_q,q)
    ={}& \sum_{c_i \in C_q} \operatorname{Rel}(c_i,q)
    - \lambda \!\!\sum_{c_i,c_j \in C_q,\, i<j}\!\! \operatorname{Red}(c_i,c_j) \\
    &+ \gamma \operatorname{Cov}(C_q,q)
    + \eta \operatorname{Conn}(C_q;\mathcal{G}),
  \end{split}
  \label{eq:context_objective}
\end{equation}
and $\mathcal{G}$ is the weighted chunk graph introduced in
Section~\ref{sec:chunk_graph}. The coefficients $\lambda,\gamma,\eta \geq 0$
control the redundancy penalty, the coverage reward, and the graph-connectivity
reward, respectively. The dense baseline of
Equation~\eqref{eq:topk} is recovered as the special case
$\lambda=\gamma=\eta=0$, in which the objective decomposes into $k$ independent
per-chunk decisions. The four terms are defined in
Section~\ref{sec:objective_terms}.

\subsection{Terms of the Context-Set Objective}
\label{sec:objective_terms}

The context-set objective is motivated by the observation that relevance alone
is insufficient for multi-hop and long-document retrieval: a selected context
A set may consist entirely of individually relevant chunks while still being
redundant, disconnected, or incomplete. Each term of
Equation~\eqref{eq:context_objective} targets one of these failure modes.

\textbf{Relevance.} $\operatorname{Rel}(c_i,q)$ is initialized from the dense
score $s_0(c_i,q)$ and updated after graph expansion, so that a chunk's
The contribution reflects both its direct similarity to the query and its position
relative to strongly relevant evidence.

\textbf{Redundancy.} $\operatorname{Red}(c_i,c_j)$ measures the semantic overlap
between two selected chunks. A highly redundant context set spends limited
context-window capacity repeating the same evidence, displacing chunks that
would otherwise contribute new information to the generator.

\textbf{Coverage.} $\operatorname{Cov}(C_q,q)$ rewards the number of distinct
evidence aspects of the query represented in the set. It is our operationalization
of \emph{evidence completeness}: whether the selected context carries enough
information to support the final answer, particularly when the reasoning chain
requires bridge evidence spanning multiple chunks or documents.

\textbf{Connectivity.} $\operatorname{Conn}(C_q;\mathcal{G})$ measures the
structural relationships among the selected chunks in the weighted chunk graph.
Chunks linked through shared entities, document adjacency, and section hierarchy,
citations, or other relational signals, are more likely to form a coherent
evidence path than an equally relevant but mutually disconnected set.

Because Equation~\eqref{eq:context_objective} is defined over sets rather than
individual chunks, it cannot be optimized by ranking alone. SpectRAG therefore
uses dense and hybrid retrieval only for high-recall candidate seeding, and then
applies graph expansion, proximity-based re-scoring, and token-budgeted
selection to approximate the maximizer of $\mathcal{F}$, yielding the final context set is passed to the generator.

\section{Spectral Graph-Augmented Retrieval}
\subsection{Overview of SpectRAG}
\label{sec:overview}

\begin{figure*}
  \centering
  \includegraphics[width=0.9\textwidth]{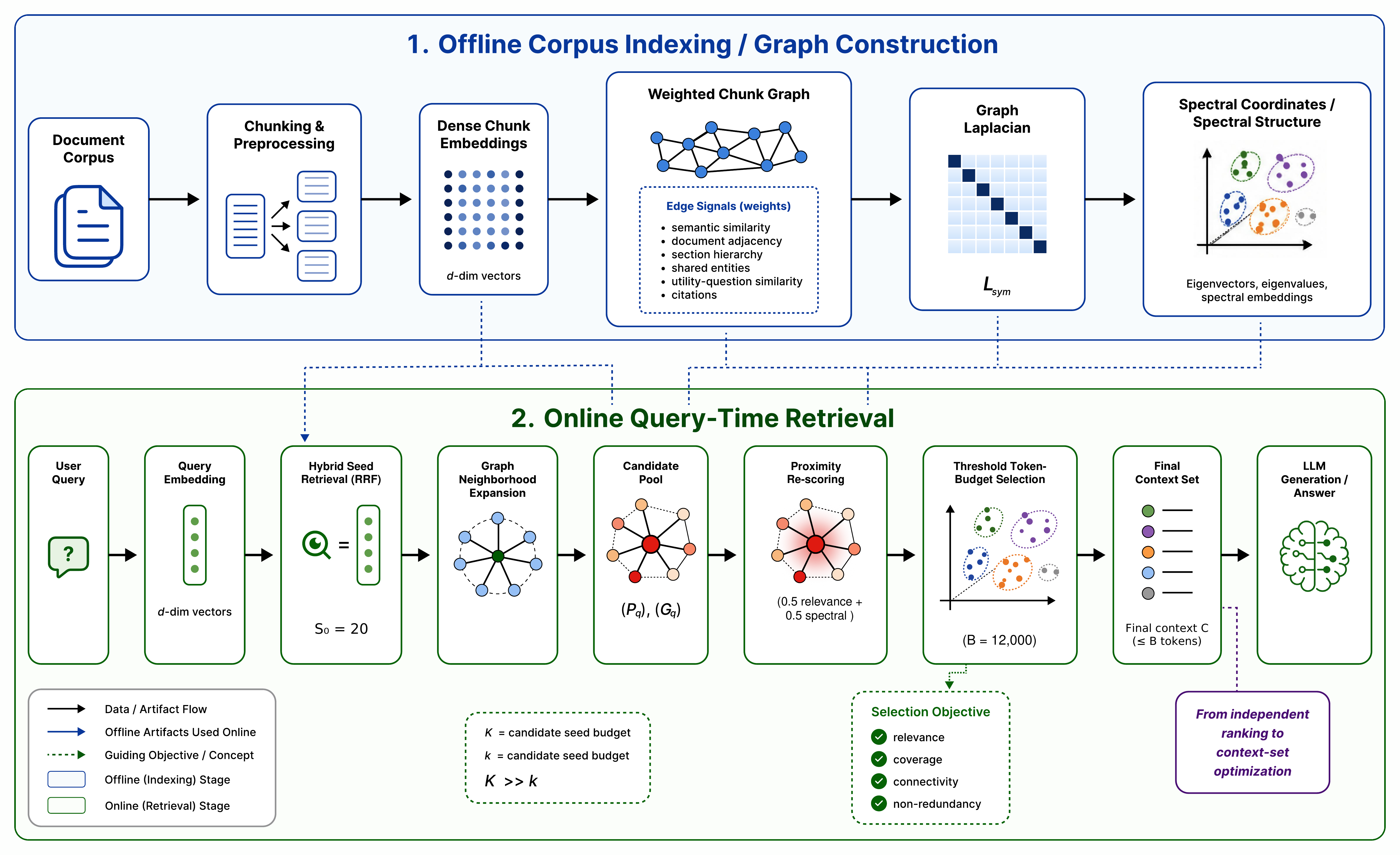}
  \caption{Overview of the SpectRAG pipeline.}
  \label{fig:spectrag_pipeline}
\end{figure*}

The offline indexing stage is summarized in Table~\ref{tab:spectrag_indexing_pipeline}. This stage is executed once for each document collection and includes token-based chunking, dense embedding computation, edge-signal extraction, weighted graph construction, spectral decomposition, Neo4j graph persistence, and disk-level storage of chunks, embeddings, spectral coordinates, utility questions, and FAISS indexes. This design separates expensive corpus-level preprocessing from query-time retrieval, allowing the online pipeline to reuse precomputed graph and spectral artifacts. Furthermore, the online query-time procedure is outlined in Table~\ref{tab:spectrag_online_pipeline}. For each query, SpectRAG first performs hybrid seed retrieval by fusing dense FAISS scores, BM25 lexical retrieval, and utility-question matching. The seed set is then expanded through two-hop graph traversal, re-scored using a combination of relevance and graph proximity, and assembled into a final context using token-budgeted selection. The final context is grouped by source document and passed to the generator together with ranked source metadata. As illustrated in Fig.~\ref{fig:spectrag_pipeline}, SpectRAG consists of two main stages: offline corpus indexing and online query-time retrieval. In the offline stage, the corpus is segmented into chunks, embedded into dense vector representations, organized into a weighted chunk graph, and transformed into graph-spectral coordinates via the Laplacian. In the online stage, SpectRAG performs dense candidate seeding, graph neighborhood expansion, spectral relevance propagation, and token-budgeted context selection before passing the final context to the generator.

\begin{table*}[t]
  \caption{Offline SpectRAG pipeline---runs once per document set}
  \label{tab:spectrag_indexing_pipeline}
  \centering
  \footnotesize
  \renewcommand{\arraystretch}{1.25}
  \begin{tabular}{@{}p{0.18\textwidth}p{0.55\textwidth}p{0.21\textwidth}@{}}
    \toprule
    \textbf{Step} & \textbf{What it does} & \textbf{Output} \\
    \midrule
    Chunking & Splits documents into structural (heading-aware) or semantic chunks measured with \texttt{tiktoken} (1200 tokens, 100 overlap). Extracts tables. Never cuts mid-sentence. & \texttt{Chunk} objects with text, position, section path, and references \\
    Embedding & Encodes each chunk with \texttt{nomic-embed-text} (768-dimensional), using the model's \texttt{search\_document:} task prefix. Builds a FAISS flat index (\texttt{IndexFlatIP}) over L2-normalized vectors for exact inner-product search. & \texttt{chunk.embedding} per chunk \& \texttt{faiss.index} on disk \\
    Edge signals & Runs six extractors to find relationships between chunks. Entity edges use \texttt{gpt-4o-mini} concept extraction by default, \& utility-question edges use \texttt{gpt-4o-mini}; both store per-chunk outputs for query-time reuse. Concepts are disk-cached. & Six lists of \texttt{RawEdge(i, j, weight)} \& \texttt{chunk. utility\_questions} \\
    Graph build & Combines all six signals using configurable weights, either fixed priors or auto-calibrated per corpus. Drops edges below the sparsification threshold. & Weighted graph $G=(V,E,W)$ \\
    Spectral decomp & Computes the normalized Laplacian $L=I-D^{-1/2}WD^{-1/2}$. Extracts the 32 smallest nontrivial eigenvectors. Each chunk gets a 32-dimensional coordinate encoding its position in the graph's community structure. & \texttt{chunk.spectral\_coords} (32-dimensional) per chunk \\
    Neo4j write & Writes all chunks as nodes and edges with per-signal attributes to Neo4j via batched Cypher. & Persistent graph in Neo4j \\
    Persist & Saves chunks, including embeddings, spectral coordinates, and utility questions, and the FAISS index to disk. & \texttt{chunks.pkl} \& \texttt{faiss. index} \\
    \bottomrule
  \end{tabular}
\end{table*}

\begin{table*}[t]
  \caption{Online SpectRAG pipeline---runs per query.}
  \label{tab:spectrag_online_pipeline}
  \centering
  \footnotesize
  \setlength{\tabcolsep}{4pt}
  \renewcommand{\arraystretch}{1.18}
  \begin{tabular}{@{}p{0.17\textwidth}p{0.59\textwidth}p{0.20\textwidth}@{}}
    \toprule
    \multicolumn{1}{c}{\textbf{Stage}} &
    \multicolumn{1}{c}{\textbf{What it does}} &
    \multicolumn{1}{c}{\textbf{Output}} \\
    \midrule
    Hybrid seed retrieval & Fuses three rankings with reciprocal rank fusion (RRF): dense FAISS cosine, utility-question matching (query to stored questions), \& BM25 lexical retrieval. The full relevance vector consumed downstream is hybrid (dense \& BM25), so lexical precision reaches later stages. & Seed set $S_0$ (20 chunks) \& hybrid relevance scores for all chunks \\
    Graph expansion & Performs breadth-first search from $S_0$ in Neo4j, using two hops and a cap of 200 nodes. Fetches the candidate subgraph. & Candidate set $S_1$ ($\leq 200$ chunks) and subgraph \\
    Proximity re-scoring & Blends relevance with graph proximity as $(1-\alpha)\cdot\mathrm{relevance}+\alpha\cdot\mathrm{proximity}$, where proximity in \texttt{hybrid} mode is the mean of spectral eigenspace closeness and Personalized PageRank from the seeds over the real weighted subgraph. & Re-scored candidate set $S_2$ \\
    Token-budget selection & Fills an approximately 12k-token budget with relevant chunks: a protected core, a tail precision floor, document-coverage tie-break, and backfill so the budget is never starved. Alternative modes include \texttt{anchor\_expand\_ce}, \texttt{protected\_mmr}, and \texttt{mmr}. & Final context set $C$ (adaptive, approximately 10--15 chunks) \\
    Generation & Formats $C$ grouped by source document and sends it to \texttt{gpt-4o-mini} with a synthesis-oriented, grounded prompt and temperature 0. & Answer string and ranked source metadata \\
    \bottomrule
  \end{tabular}
\end{table*}

\subsection{Chunk Graph Construction}
\label{sec:chunk_graph}

The context-set objective in Equation~\eqref{eq:context_objective} is defined
with respect to a weighted chunk graph $\mathcal{G}$: the connectivity term
$\operatorname{Conn}(C_q;\mathcal{G})$ is evaluated on its edges, and the
expansion and re-scoring stages of Section~\ref{sec:overview} operate on its
spectrum. This subsection specifies how $\mathcal{G}$ is constructed, weighted,
sparsified, and decomposed during offline indexing.

\subsubsection{Nodes and edge signals}
The chunks produced by
the segmentation stage, each carrying its dense embedding
$\mathbf{e}_i \in \mathbb{R}^{d}$. Unlike a $k$-nearest-neighbour embedding
graph, edges are drawn from six complementary evidence signals, each normalized
to $[0,1]$:

\begin{itemize}
  \item \textbf{Semantic} ($w_{ij}^{sem}$): truncated cosine similarity between
    embeddings, $\max(0,\cos(\mathbf{e}_i,\mathbf{e}_j))$, restricted to the
    $k_g$ nearest neighbours of each node.
  \item \textbf{Adjacency} ($w_{ij}^{adj}$): consecutive chunks within the same
    source document, capturing narrative continuity across chunk boundaries.
  \item \textbf{Hierarchy} ($w_{ij}^{sec}$): shared headings or a common parent
    section in the document's structural path.
  \item \textbf{Entity} ($w_{ij}^{ent}$): overlap in named entities, technical
    terms, organizations, or dates extracted per chunk.
  \item \textbf{Utility} ($w_{ij}^{utl}$): similarity between LLM-generated
    questions that each chunk answers, encoding functional rather than
    surface similarity~\cite{rappazzo2024gemrag}.
  \item \textbf{Citation} ($w_{ij}^{cit}$): explicit references, hyperlinks,
    footnotes, or dependency relations between chunks.
\end{itemize}

\subsubsection{Weight fusion and sparsification}
The six signals are combined into a single edge weight,
\begin{equation}
  w_{ij} =
  \alpha_1 w_{ij}^{sem} + \alpha_2 w_{ij}^{adj} + \alpha_3 w_{ij}^{sec}
  + \alpha_4 w_{ij}^{ent} + \alpha_5 w_{ij}^{utl} + \alpha_6 w_{ij}^{cit},
  \label{eq:edge_weight}
\end{equation}
with $\alpha_r \ge 0$ set either from fixed priors or auto-calibrated per corpus.
Edges with $w_{ij}<\tau$ are discarded, keeping $|E|=O(k_g n)$ so that the
subsequent traversal and spectral steps remain tractable at corpus scale.

Purely semantic edges reproduce the redundancy problem of
Section~\ref{sec:limitations}: near-duplicate chunks form tight cliques while
bridge evidence stays disconnected. The non-semantic signals in
Equation~\eqref{eq:edge_weight} are precisely those that link chunks a dense
retriever scores independently, and they supply the paths along which relevance
is later diffused.

\subsubsection{Spectral coordinates}
Let $W$ be the resulting weighted adjacency matrix and $D$ the diagonal degree
matrix, $D_{ii}=\sum_j w_{ij}$. SpectRAG uses the symmetric normalized
Laplacian
\begin{equation}
  L_{\mathrm{sym}} = I - D^{-1/2} W D^{-1/2},
  \label{eq:laplacian}
\end{equation}
whose $m$ smallest non-trivial eigenvectors, stacked columnwise as
$U \in \mathbb{R}^{n \times m}$, assign each chunk a spectral coordinate
$\mathbf{u}_i \in \mathbb{R}^{m}$ ($m=32$ in our implementation). Because the
eigenvectors of $L_{\mathrm{sym}}$ minimize a normalized-cut relaxation, chunks
within the same latent community are close in $\mathbb{R}^{m}$ while chunks
separated by a sparse cut are distant, irrespective of their raw embedding
similarity. Proximity in this space therefore provides a structural
complement to $s_0(c_i,q)$ at query time, and $L_{\mathrm{sym}}$ additionally
serves as the diffusion operator for relevance propagation.

The construction is illustrated in Fig.~\ref{fig:weighted_chunk_graph}, and
is performed once per corpus; only $U$, $W$, and the chunk embeddings are
persisted for online retrieval.

\subsection{Documents Chunking}
The chunking process involves adaptive structure-aware chunking, which is done by picking one of two segmentations for each document. Documents with explicit structure information, such as Markdown markup or section numbering, employ the structural segmentation, splitting text along the hierarchical structure, propagating the entire path through the sections hierarchy (e.g., 2. Methods → 2.1 Datasets) as chunk metadata; the section-affinity edge generation will use this hierarchy. For unstructured text, the semantic mode is applied: sentences are embedded using the same model as for document embedding (nomic-embed-text), and breakpoints are inserted at places where the embedding distance between two sentences exceeds the 90th percentile of the entire document distance distribution, resulting in topically coherent segments independently of the document surface format. In both segmentation modes, the chunk size is measured in tokens using the \texttt{cl100k\_base} tokenizer, with a maximum number of tokens set to 1,200 tokens with 100-token overlap as a suffix of the previous chunk that respects sentence boundaries. Chunks with less than 32 tokens are concatenated with the previous chunk, and tables are parsed separately, serializing them to Markdown so that rows are not split.

We selected a granularity of 1,200 tokens for analysis. The decision to increase the token size from 256 to 1,200 resulted in the generation of more coherent text segments, which significantly enhanced the efficacy of abstractive sensemaking queries. Consequently, this modification contributed to an improvement in the overall quality of information retrieval.

\subsection{Laplacian Spectral Coordinates}
\begin{figure}
  \centering
  \includegraphics[width=0.5\textwidth]{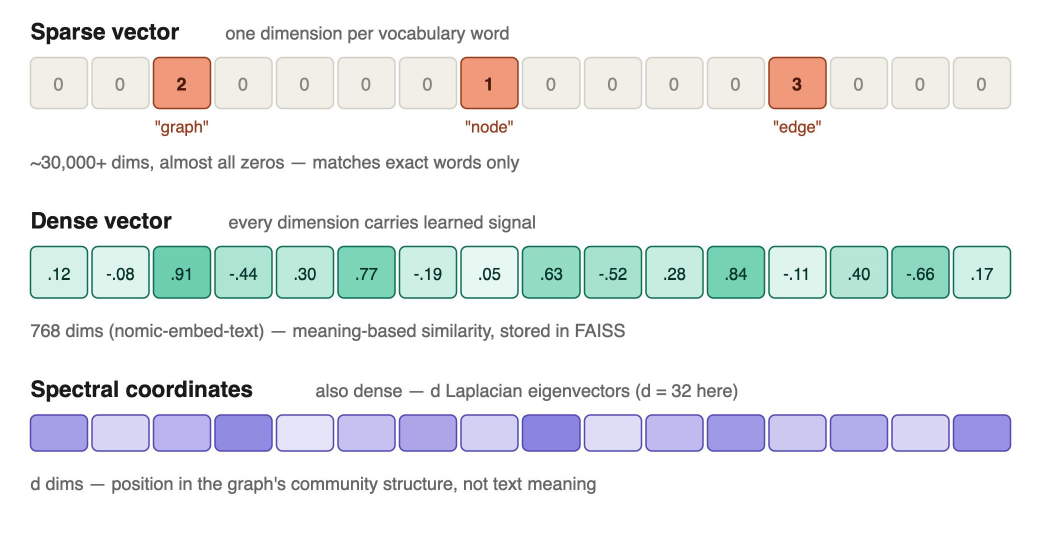}
  \caption{Sparse lexical vectors (top) allocate one dimension per vocabulary term and match exact words; dense embeddings (middle, 768-d) encode meaning in every dimension and support similarity search via FAISS.  Bottom: SpectRAG's spectral coordinates ($d=32$), derived from the
  smallest eigenvectors of the normalized graph Laplacian, encode a chunk's position in the corpus graph's community structure rather than its textual semantics.}
  \label{fig:vector_representations}
\end{figure}

SpectRAG assigns each chunk $d$-dimensional spectral coordinates (bottom, $d=32$), derived from the normalized graph Laplacian, encoding position in the corpus graph's community structure rather than textual semantics. As shown in Fig.~\ref{fig:vector_representations}, the representation is compared with typical vectors used in retrieval. Despite being similar in structure to the dense embedding, a compact, continuous vector used for cosine similarity, spectral coordinates carry an entirely different kind of signal. While $e_i$ holds information about the content of a chunk, $z_i$ contains information about its location within the weighted chunk graph. This means that chunks from the same community in the chunk graph have close coordinates even if they do not match in their surface text. It is crucial to understand that SpectRAG does not depend on sparse retrieval; the sparse vector appears in Fig.~\ref{fig:vector_representations} only for contrast. Section~\ref{sec:spectral-diffusion} shows how the two complementary geometries of Fig.~\ref{fig:vector_representations} (middle vs.\ bottom) are combined: dense embedding similarity supplies the initial relevance signal, which is then diffused over the chunk graph whose low-frequency structure the spectral coordinates $z_i$ encode.

We compute the first $m$ nontrivial eigenvectors of $L_{sym}$:
\begin{equation}
  L_{sym}U = U\Lambda.
\end{equation}
The spectral coordinate of chunk $c_i$ is
\begin{equation}
  z_i = [u_2(i), u_3(i), \ldots, u_{m+1}(i)] \in \mathbb{R}^{m}.
\end{equation}
These coordinates capture low-frequency variation over the graph. Chunks that are structurally close, or that occupy similar positions in graph communities, tend to be close in spectral space.

\subsection{Initial Dense Retrieval}
At query time, SpectRAG first forms a seed set $S_0$ from three rankings: dense FAISS similarity, BM25 lexical scores, and similarity between the query and the stored utility questions. The rankings are fused with RRF, and the top $K = 20$ chunks are kept (Table~\ref{tab:spectrag_online_pipeline}). This stage only decides where graph expansion starts; the final context is selected later by graph expansion, proximity re-scoring, and token-budgeted selection. $K$ is therefore set larger than the expected size of the final context, which is about 10--15 chunks under the 12k-token budget:

\begin{equation}
  K \gg k.
\end{equation}
The final context set is selected later by graph expansion, spectral diffusion, and token-budget final context selection.

\subsection{Graph Expansion}
The dense retrieval approach finds chunks which individually resemble the query; whereas in multi-hop reasoning, evidence may depend on information that is structurally, but not semantically, related to the query. These chunks, by their very nature, will be outside the scope of nearest neighbor search as they are situated outside the embedding neighborhood of the query and still have structural connections to the query via the graph $G$ .

\paragraph{Seed set.}
Let $S_0 \subset V$ denote the seed set returned by hybrid retrieval,

\begin{equation}
  \begin{aligned}
    S_0 &= \operatorname*{Top\text{-}K}_{c_i \in V} \mathrm{RRF}(c_i, q), \qquad |S_0| = K,\\
    \mathrm{RRF}(c_i, q) &= \sum_{r \in \mathcal{R}} \frac{1}{\kappa + \mathrm{rank}_r(c_i)},
  \end{aligned}
\end{equation}
where $\mathcal{R}$ contains the dense, BM25, and utility-question rankings, $\mathrm{rank}_r(c_i)$ is the position of $c_i$ in ranking $r$, $\kappa$ is the RRF constant, and $K$ is the candidate budget. Over-retrieving at this stage is deliberate: expansion can only recover evidence reachable from the seeds, so recall at $S_0$ upper-bounds recall of the entire pipeline.

\subsection{Spectral Diffusion of Query Relevance}
\label{sec:spectral-diffusion}

Given an initial relevance vector $s \in \mathbb{R}^n$ with
\begin{equation}
  s_i = \operatorname{sim}(e_q,e_i),
\end{equation}
we define a diffused relevance signal:
\begin{equation}
  s' = \exp(-tL_{sym})s,
  \label{eq:heat_diffusion}
\end{equation}
where $t$ controls diffusion scale. Small $t$ keeps relevance local; large $t$ spreads relevance across broader graph regions.

For large corpora, Equation~\ref{eq:heat_diffusion} can be approximated by truncated Chebyshev polynomials, personalized PageRank-style iteration, or repeated smoothing:
\begin{equation}
  s^{(r+1)} = \beta s + (1-\beta)\tilde{A}s^{(r)},
\end{equation}
where $\tilde{A}=D^{-1/2}AD^{-1/2}$ and $\beta \in (0,1)$.

\subsection{Token-Budgeted and Cluster-Aware Context Selection}

After candidate expansion and proximity-based re-scoring, SpectRAG constructs the final context set under a fixed token budget rather than selecting a fixed number of chunks. This design is motivated by the fact that retrieval-augmented generation systems are constrained by the input context window of the generator. Therefore, the final retrieval stage should optimize not only which chunks are selected, but also how the available token budget is allocated across evidence sources.

Let $\mathcal{P}q$ denote the re-scored candidate pool for query $q$, and let $\ell(c_i)$ be the token length of candidate chunk $c_i$. Given a maximum context budget $B$, the final context set $C_q$ is selected subject to:
\begin{equation}
  \sum{c_i \in C_q} \ell(c_i) \leq B.
\end{equation}
The token-budgeted selection objective can be written as:
\begin{equation}
  C_q^{*}
  =
  \arg\max_{C_q \subseteq \mathcal{P}q}
  \mathcal{F}(C_q,q)
  \quad
  \mathrm{s.t.}
  \quad
  \sum{c_i \in C_q} \ell(c_i) \leq B,
\end{equation}
where $\mathcal{F}(C_q,q)$ denotes the utility of the selected context set.

In practice, SpectRAG approximates this budget-constrained objective using a token-aware selection strategy. At each step, the candidate score combines relevance, graph-based proximity, spectral-cluster coverage, document coverage, and redundancy control:
\begin{equation}
  \begin{aligned}
    U(c_i \mid C_q) ={}& a \cdot s_i
    + b \cdot \operatorname{Prox}(c_i,S_0) \\
    &+ c \cdot \operatorname{ClusterGain}(c_i,C_q) \\
    &+ d \cdot \operatorname{DocGain}(c_i,C_q) \\
    &- \lambda \cdot \operatorname{Red}(c_i,C_q).
  \end{aligned}
\end{equation}
where $s_i$ is the re-scored relevance value of candidate $c_i$, $\operatorname{Prox}(c_i,S_0)$ measures graph proximity to the seed set $S_0$, $\operatorname{ClusterGain}(c_i,C_q)$ rewards coverage of underrepresented spectral clusters, $\operatorname{DocGain}(c_i,C_q)$ encourages coverage across source documents, and $\operatorname{Red}(c_i,C_q)$ penalizes semantic overlap with already selected chunks.
To account for chunk length, candidates may be selected according to token-normalized utility:
\begin{equation}
  c^{*}
  =
  \arg\max_{c_i \in \mathcal{P}_q \setminus C_q}
  \frac{U(c_i \mid C_q)}{\ell(c_i)^{\rho}},
\end{equation}
where $\rho \in [0,1]$ controls the strength of length normalization. When $\rho=0$, selection is based only on utility; when $\rho=1$, the method favors candidates with higher utility per token.

The selection procedure consists of three practical components. First, SpectRAG preserves a protected core of high-confidence chunks so that strongly relevant evidence is not removed by later diversity or budget constraints. Second, it applies a precision-oriented tail selection strategy for lower-ranked candidates, using graph proximity, cluster coverage, document coverage, and redundancy penalties to decide which additional chunks should enter the context. Third, if the budget is not fully used, a backfill step adds remaining high-quality candidates so that the generator is not deprived of useful evidence.

This token-budgeted design differs from fixed-$k$ context selection. Instead of forcing every query to receive the same number of chunks, SpectRAG adapts the number of selected chunks to their token lengths and the available context window. In the current implementation, this typically produces a final context of approximately 10--15 chunks under a context budget of about 12k tokens. The resulting context set is intended to be relevant, non-redundant, structurally connected, and sufficiently broad for grounded generation.

\begin{algorithm}
  \caption{Online \textit{SpectRAG} retrieval and generation.}
  \label{alg:SpectRAG}
  \begin{algorithmic}[1]
    \REQUIRE Query $q$, chunks $V$, FAISS index, BM25 index, utility questions, graph $G=(V,E,W)$, spectral coordinates $Z$, token budget $B$
    \STATE Embed $q$ as $e_q$ and compute dense FAISS scores
    \STATE Compute BM25 scores and utility-question matching scores
    \STATE Fuse rankings with RRF to form seed set $S_0$
    \STATE Fetch candidate subgraph by two-hop BFS from $S_0$ in Neo4j
    \STATE Let $S_1$ be the expanded candidate set, capped at 200 chunks
    \STATE Compute spectral proximity from $Z$ and Personalized PageRank on the candidate subgraph
    \STATE Re-score candidates: $s_2(c) \gets (1-\alpha)r(c)+\alpha p(c)$ for $c \in S_1$
    \STATE Initialize protected core $C$ from highest-scoring candidates
    \WHILE{token count of $C$ is below $B$ and candidates remain}
    \STATE Add the best candidate satisfying the precision floor and coverage tie-breaks
    \ENDWHILE
    \STATE Backfill with highest-scoring remaining chunks if budget remains
    \STATE Format $C$ grouped by source document and generate answer with grounded prompt
    \STATE \RETURN Answer, ranked source metadata, and final context set $C$
  \end{algorithmic}
\end{algorithm}

\section{Experiments}
The purpose of the experiments is to test whether SpectRAG improves retrieval and generation quality when evidence is distributed, multi-hop, or structurally connected. We compare SpectRAG against graph-based and vector-based RAG baselines along four dimensions: index cost, holistic answer quality, answer accuracy, and evidence ranking.

\subsection{Research Questions}
\begin{itemize}
    \setlength{\itemsep}{0pt}
    \setlength{\parskip}{0pt}
    \setlength{\parsep}{0pt}
  \item \textbf{RQ1:} Does SpectRAG's chunk-level graph achieve competitive retrieval at a lower index cost than LightRAG's entity-level graph?
  \item \textbf{RQ2:} Does SpectRAG improve holistic answer quality (comprehensiveness, diversity, empowerment) over graph-RAG baselines?
  \item \textbf{RQ3:} Does SpectRAG improve objective answer accuracy (Exact Match ($EM$), $F1$) on multi-hop questions with gold answers, compared to graph- and vector-RAG baselines?
  \item \textbf{RQ4:} Does SpectRAG rank gold evidence higher in the retrieved context than graph- and vector-RAG baselines, not merely retrieve more of it?
\end{itemize}

\subsection{Evaluation Datasets} % - UltraDomain - done
To estimate SpectRAG's effectiveness across a range of domains and content types, we used four datasets from the UltraDomain \cite{b16} and employed specific evaluation metrics such as precision, recall and f1 to quantify performance. Additionally, to evaluate against hallucination, we added the MultiHopRAG \cite{b17} dataset, which tests the model's ability to avoid generating unsupported information.

The UltraDomain benchmark is derived from 428 college textbooks covering 18 domains, including agriculture, the social sciences, and the humanities \cite{b16}. We selected the agriculture, computer science (CS), legal, and mixed (Mix) domains to focus on key areas relevant to our research, each with a token size of 600,000-5,000,000. Four domains used for our experiment are listed below:
\begin{itemize}
  \item \textbf{Agriculture:} This field emphasizes agricultural methods, addressing subjects like beekeeping, hive supervision, crop cultivation, and disease management \cite{b17}.
  \item \textbf{CS:} This particular subset relates to computer science books on algorithms, systems, and programming principles. Similar to agriculture, it serves as an example in the subject because the questions and texts share a common language, even from different textbooks \cite{b16}.
  \item \textbf{Legal:} This domain focuses on legal practices in corporate settings, including case law, statutes, contracts, regulatory documents, and governance, emphasizing the legal and financial industries \cite{b16, b17}.
  \item \textbf{Mixed:} This subset collected from multiple domain such as literary, philosophical texts, historical studies and other writing styles \cite{b16, b17}.
\end{itemize}

\subsection{Implementation Parameters and Details}
For the analysis of the impact of retrieval granularity, we consider SpectRAG under two token-level chunking schemes. In the first token-based configuration, the chunk size is set to 256 tokens with an overlap of 64 tokens. This setting produces smaller retrieval units and allows the token-budgeted selector to compose evidence at a finer granularity. The second scheme utilizes a chunk size of 1200 tokens with an overlap of 100 tokens. This scheme produces larger retrieval chunks and enables the rich local context of documents to be retained in the retrieved candidates.

The implementation parameters used in the SpectRAG pipeline are reported in Table~\ref{tab:implementation_parameters}. These parameters include the token-based chunking configuration, embedding model, edge extraction modules, LLM-assisted extraction components, graph and spectral index settings, seed retrieval strategy, graph expansion radius, proximity re-scoring method, and token-budgeted context selection strategy. Reporting these settings is important for reproducibility because SpectRAG depends on both offline graph construction choices and online retrieval-stage parameters.

The two configurations above share a common retrieval pipeline consisting of hybrid candidate generation, graph neighborhood expansion, proximity-based re-scoring, and token-budgeted final context selection.
\begin{table*}[h]
  \caption{Implementation parameters aligned with the SpectRAG pipelines}
  \label{tab:implementation_parameters}
  \centering
  \footnotesize
  \renewcommand{\arraystretch}{1.25}
  \begin{tabular*}{\textwidth}{@{\extracolsep{\fill}}p{0.22\textwidth}p{0.50\textwidth}p{0.20\textwidth}@{}}
    \toprule
    \textbf{Parameter} & \textbf{Setting} & \textbf{Related stage} \\
    \midrule
    Chunking & Structural or semantic chunking with \texttt{tiktoken}; 1200-token chunks and 100-token overlap; sentence boundaries and tables are preserved. & Offline chunking \\
    Embedding model & \texttt{nomic-embed-text}, 768-dimensional embeddings, with the \texttt{search\_document:} prefix for indexed chunks. & Offline embedding \\
    Edge extraction & Six edge-signal extractors: semantic, document adjacency, section hierarchy, entity/concept, utility-question, and citation/dependency signals. & Offline edge signals \\
    LLM-assisted extraction & \texttt{gpt-4o-mini} for concept extraction, utility-question generation, and answer generation; \texttt{gpt-4o} for judging when evaluation requires an LLM judge. & Offline and generation \\
    Graph and spectral index & Weighted graph $G=(V,E,W)$ with sparsified edges; normalized Laplacian $L=I-D^{-1/2}WD^{-1/2}$; 32 spectral coordinates per chunk. & Graph build and spectral decomp \\
    Seed retrieval & RRF over dense FAISS cosine, BM25 lexical retrieval, and utility-question matching; seed set size is 20 chunks. & Online stage 1 \\
    Graph expansion & Two-hop BFS from $S_0$ in Neo4j, capped at 200 candidate chunks. & Online stage 2 \\
    Re-scoring & Hybrid proximity blends relevance with spectral eigenspace closeness and Personalized PageRank over the weighted subgraph. & Online stage 3 \\
    Context selection & Approximately 12k-token budget with protected core, tail precision floor, document-coverage tie-breaks, and backfill; ablations include \texttt{anchor\_expand\_ce}, \texttt{protected\_mmr}, and \texttt{mmr}. & Online stage 4 \\
    \bottomrule
  \end{tabular*}
\end{table*}

\subsection{Baselines}
We compare SpectRAG against the following state-of-the-art approaches:
\begin{itemize}
  \item \textbf{NaiveRAG} \cite{b19}: This model follows the conventional indexing, retrieval, and generation process. Chunks are extracted from documents, embedded, and stored in a vector database; during inference, the query is embedded, and the top-$k$ most similar chunks are retrieved and concatenated into a prompt for an LLM to generate the answer.

  \item \textbf{HybridRAG} \cite{b20}: This approach integrates BM25 sparse lexical retrieval and dense semantic retrieval, merging their ranked lists using RRF. In addition, uncertainty quantification (UQ) is incorporated to estimate the confidence of the retrieved evidence and the generated response, allowing low-reliability outputs to be flagged.

  \item \textbf{LightRAG} \cite{b18}: This model uses an LLM to extract entities and relations from documents and stores them as a knowledge graph, as implemented in LightRAG. Retrieval is performed at two levels, namely entity-level and thematic-level, enabling the system to capture multi-hop and cross-document dependencies that flat chunk-based indexes cannot represent.

  \item \textbf{HyDeRAG} \cite{b21}: Retrieval based on Hypothetical Document Embeddings (HyDE), which closes the semantic gap between a short query and a document by using an LLM to generate a hypothetical answer to the query. The generated synthetic document is then embedded and used in place of the raw query embedding for the similarity search, increasing the likelihood of retrieving the relevant chunks.
\end{itemize}

\subsection{Retrieval and Answer Correctness Metrics}
\label{sec:pr-metrics}
The measures of Section~\ref{sec:eval-metrics} are \emph{relative}: they express a judge's preference between two systems, which helps researchers see comparative strengths. We complement them with \emph{absolute} measures against benchmark ground truth, providing a complete assessment. Reporting correctness at two levels shows the audience that different failure modes matter and need tailored solutions.

\subsubsection{Retrieval-Level Correctness}
Here, we use $C_q$ to refer to the set of contexts obtained from query $q$, and $E_q^{*}$ to be the set of evidence chunks annotated in the benchmark data. The precision will refer to the ratio of the supporting evidence contained in the retrieval results, whereas the recall will be the ratio of retrieved evidence chunks.

\begin{equation}
  \mathrm{Precision@}k \;=\; \frac{|C_q \cap E_q^{*}|}{|C_q|},
  \qquad
  \mathrm{Recall@}k \;=\; \frac{|C_q \cap E_q^{*}|}{|E_q^{*}|}.
\end{equation}
Their harmonic mean gives the balanced score
\begin{equation}
  \mathrm{F1@}k \;=\;
  \frac{2 \cdot \mathrm{Precision@}k \cdot \mathrm{Recall@}k}
  {\mathrm{Precision@}k + \mathrm{Recall@}k}.
\end{equation}

with $\mathrm{F1@}k \triangleq 0$ when both terms vanish. All three are averaged over the query set. Because $|C_q| = k$ is fixed by the context budget, $\mathrm{Precision@}k$ and $\mathrm{Recall@}k$ are not independent: enlarging $k$ monotonically increases recall while generally depressing precision. We therefore report $Recall@k$ at $k$ = 2, 5 and 10 and $MRR$ for every system (Table~\ref{tab:retrieval-results}). Overall recall, precision and hit rate are computed on each system’s full retrieved context, whose size differs between systems (about 10–15 chunks for SpectRAG under the 12k-token budget), so they are not matched-budget measures.

\subsubsection{Answer-Level Correctness}
\label{sec:eval-metrics}
In case of benchmark test cases where the gold-standard short answers are known, the standard extractive QA approach is followed. Let $\hat{a}$ be the output answer, and $a^{*}$ be the reference answer; both have been normalized using lower casing and stripping off articles and punctuation.

\begin{equation}
  \mathrm{EM} \;=\; \mathbf{1}\bigl[\, \hat{a} = a^{*} \,\bigr].
\end{equation}
and token-level $F1$ is computed over the multiset of tokens
$T(\cdot)$:
\begin{equation}
  P_{\mathrm{tok}} = \frac{|T(\hat{a}) \cap T(a^{*})|}{|T(\hat{a})|},
  \qquad
  R_{\mathrm{tok}} = \frac{|T(\hat{a}) \cap T(a^{*})|}{|T(a^{*})|},
\end{equation}
\begin{equation}
  \mathrm{F1}_{\mathrm{tok}} \;=\;
  \frac{2 \cdot P_{\mathrm{tok}} \cdot R_{\mathrm{tok}}}
  {P_{\mathrm{tok}} + R_{\mathrm{tok}}}.
\end{equation}

Where a question admits several references, we take the maximum over references, as is standard.

Furthermore, we use to judge the answers that the model produces, these are the four measures that:
\begin{itemize}
  \item \textbf{Comprehensiveness}: It is a measure of the ability of the answer to provide an account of all facets of the question.
  \item \textbf{Diversity}: It is a measure of the diversity of viewpoints in the answer relative to the question.
  \item \textbf{Empowerment}: It is a measure of how much an answer helps a reader understand and make judgments on a topic.
  \item \textbf{Overall}: This is a composite score of the first three measures.
\end{itemize}

\subsubsection{Applicability}
These measures require labeled ground truth and are therefore reported only on MultiHop-RAG and UltraDomain,  which annotate gold supporting facts and short answers. The sensemaking protocol of Section~\ref{sec:eval-metrics} produces open-ended questions without references and admits only the preference-based measures. The two families are reported separately and are not aggregated: they quantify different properties, and a system may improve evidence recall without improving preference, or vice versa.

\subsection{Preliminary Generation-Level Evaluation}
For the preliminary generation-level evaluation, we compared SpectRAG and NaiveRAG on a subset of the UltraDomain dataset using a pairwise approach. For each domain, we indexed all documents and selected 20 sample questions. An LLM-as-judge protocol was used, with the judge model kept distinct from the generation model to eliminate self-preference bias. The judge received both candidates' answers and issued the final verdict, operating at 0 temperature to maximize determinism. The evaluation follows a LightRAG-style sensemaking protocol, in which answers are judged pairwise across multiple qualitative dimensions. To mitigate position bias, the answer order was swapped during judging.
The pairwise win rates of SpectRAG against each baseline are reported in Tables~\ref{tab:winrates-config1} and~\ref{tab:winrates-config2}, where the baseline and SpectRAG values in each column pair sum to 100\%. Values above 50\% indicate an advantage for SpectRAG over NaiveRAG. Averaged over the four domains, SpectRAG outperformed NaiveRAG on all four dimensions under both signal configurations, with the largest margin observed for diversity (64.1\% in Table~\ref{tab:winrates-config1} and 60.0\% in Table~\ref{tab:winrates-config2}). The exception is the Legal domain, where NaiveRAG is slightly preferred on every dimension.
\subsection{RQ1}
\label{sec:rq1}

Because SpectRAG defines its graph over document chunks rather than LLM-extracted entities, it represents every evaluated corpus with substantially fewer nodes and a smaller on-disk index, while matching or exceeding LightRAG on retrieval quality. The node reduction holds across all five corpora, ranging from $2.8\times$ to $6.6\times$ fewer nodes, and the complete persisted index on MultiHopRAG is $6\times$ smaller (46~MB versus 274~MB), since SpectRAG maintains only a single chunk-level vector index and a spectral basis and eliminates LightRAG's separate entity (100~MB) and relationship (106~MB) vector databases. This lighter representation does not sacrifice quality: on MultiHopRAG's gold-evidence annotations, SpectRAG attains a $Recall@5$ of 64.7 against LightRAG's 49.2, with comparable answer accuracy. SpectRAG therefore realizes a favorable cost and quality trade-off, offering a compact, denser chunk graph that competes with a state-of-the-art entity-graph baseline at a fraction of its index footprint. Table~\ref{tab:rq1_graph_size} reports the per-corpus graph size and index cost for both systems. In addition, the results support four key findings:

\begin{enumerate}
  \item \textbf{Roughly three to seven times fewer nodes across every corpus ($2.8\times$ to $6.6\times$).} SpectRAG never materializes LightRAG's approximately 15{,}000 entity nodes, so its spectral and personalized PageRank (PPR) diffusion runs over a far smaller node set.
  \item \textbf{Approximately six times smaller index (46~MB versus 274~MB on MultiHopRAG).} The gap is almost entirely attributable to LightRAG's entity and relationship vector stores (206~MB combined), which SpectRAG does not build.
  \item \textbf{Fewer nodes and richer edges, not fewer edges.} SpectRAG's chunk graph is consistently denser per node (edges per node up to 5.7 versus LightRAG's approximately 1), so the efficiency claim is correctly stated in terms of node count and storage, never edge count.
  \item \textbf{No quality penalty.} SpectRAG achieves a $Recall@5$ of 64.7 versus 49.2 for LightRAG on MultiHopRAG, with comparable answer accuracy.
\end{enumerate}

\begin{table}[!t]
  \caption{Per-corpus graph size and index cost.}
  \label{tab:rq1_graph_size}
  \centering
  \scriptsize
  \renewcommand{\arraystretch}{1.2}
  \setlength{\tabcolsep}{3pt}
  \begin{tabular*}{\columnwidth}{@{\extracolsep{\fill}}llrrrrr@{}}
    \toprule
    \textbf{Corpus} & \textbf{System} & \textbf{Nodes} & \textbf{Edges} & \textbf{E/N} & \textbf{Graph} & \textbf{Node} \\
    & & & & & \textbf{File} & \textbf{Red.} \\
    \midrule
    Agriculture & LightRAG & 1{,}236 & 1{,}083 & 0.88 & 1.08 MB & --- \\
    & \textbf{SpectRAG} & \textbf{239} & \textbf{476} & \textbf{1.99} & \textbf{0.14 MB} & \textbf{\boldmath$5.2\times$} \\
    \addlinespace
    CS & LightRAG & 1{,}327 & 1{,}036 & 0.78 & 1.06 MB & --- \\
    & \textbf{SpectRAG} & \textbf{461} & \textbf{2{,}608} & \textbf{5.66} & \textbf{0.72 MB} & \textbf{\boldmath$2.9\times$} \\
    \addlinespace
    Legal & LightRAG & 5{,}615 & 6{,}784 & 1.21 & 6.80 MB & --- \\
    & \textbf{SpectRAG} & \textbf{1{,}989} & \textbf{10{,}019} & \textbf{5.04} & \textbf{2.76 MB} & \textbf{\boldmath$2.8\times$} \\
    \addlinespace
    Mix & LightRAG & 7{,}428 & 5{,}865 & 0.79 & 5.92 MB & --- \\
    & \textbf{SpectRAG} & \textbf{1{,}127} & \textbf{1{,}603} & \textbf{1.42} & \textbf{0.46 MB} & \textbf{\boldmath$6.6\times$} \\
    \addlinespace
    MultiHopRAG & LightRAG & 15{,}803 & 16{,}911 & 1.07 & 16.40 MB & --- \\
    & \textbf{SpectRAG} & \textbf{5{,}022} & \textbf{19{,}326} & \textbf{3.85} & \textbf{5.34 MB} & \textbf{\boldmath$3.1\times$} \\
    \bottomrule
  \end{tabular*}

  \vspace{2pt}
  \raggedright\footnotesize{Nodes are document chunks for SpectRAG and LLM-extracted entities for LightRAG. E/N is edges per node. Node reduction (Node Red.) equals LightRAG nodes divided by SpectRAG nodes. Full persisted-index footprint on MultiHopRAG: SpectRAG 46 MB versus LightRAG 274 MB (excluding LightRAG's 63 MB build-time LLM cache).}
\end{table}

\subsection{RQ2}
Partly. Against LightRAG, the result depends on the signal configuration. Averaged over the four domains, SpectRAG is preferred on diversity under both configurations (55.0\% and 55.9\%), but on the overall verdict it is behind under configuration 1 (48.7\%) and slightly ahead under configuration 2 (51.2\%) (Tables~\ref{tab:winrates-config1} and ~\ref{tab:winrates-config2}). LightRAG is preferred most clearly in the CS domain (overall 58.8\% and 55.0\%). Table~\ref{tab:cs_case_study} reveals one CS question on which SpectRAG was preferred on diversity, empowerment and the overall verdict in both answer orders, and on comprehensiveness in one order. We include it to show what the judge rewarded when SpectRAG won, not as a typical case.

\begin{table*}[!h]
  \caption{CS-domain qualitative case study: query, answers, and LLM-judge verdict.}
  \label{tab:cs_case_study}
  \centering
  \footnotesize
  \renewcommand{\arraystretch}{1.25}
  \setlength{\tabcolsep}{4pt}
  \begin{tabular*}{\textwidth}{@{\extracolsep{\fill}}p{0.16\textwidth}p{0.76\textwidth}@{}}
    \toprule
    \textbf{Field} & \textbf{Content} \\
    \midrule
    Query & How can anomaly detection techniques be applied in various domains to identify outliers? \\
    SpectRAG answer & Frames anomaly detection as a versatile toolset applied through several themes: understanding the dataset (categorical, ordinal, or continuous) to guide method choice; statistical methods (z-scores, p-values, standard-deviation thresholds); machine-learning approaches (K-means clustering, supervised learning on labeled data); domain-specific applications in finance (fraud), healthcare (rare conditions), network security (intrusions), and manufacturing (predictive maintenance); evaluation and validation via precision, recall, $F1$, and cross-validation; and visualization tools (scatter plots, box plots, heat maps) for inspecting distributions. \\
    LightRAG answer & Synthesizes anomaly detection across fields through big data and machine learning (real-time pattern flagging for finance and network security); statistical analysis (z-scores, box plots, interquartile range for healthcare); data mining (clustering to surface outliers in marketing); optimization methods (refining operations-research models); and real-time applications (streaming detection in telecommunications). Concludes that the techniques are versatile across finance, healthcare, marketing, and operations research. \\
    Comprehensiveness & \textbf{Winner: SpectRAG} (position-sensitive; split on swap). More detailed exploration covering a wider range of methods, applications, and evaluation strategies. \\
    Diversity & \textbf{Winner: SpectRAG} (confirmed both orderings). Richer variety of perspectives across multiple domains and specific techniques within each. \\
    Empowerment & \textbf{Winner: SpectRAG} (confirmed both orderings). Detailed explanations and examples that help the reader understand how the techniques are applied. \\
    Overall & \textbf{Winner: SpectRAG} (confirmed both orderings). Superior comprehensive coverage, diverse insights, and deeper reader understanding of the topic. \\
    Analysis & SpectRAG wins all three quality dimensions and the overall verdict. Three of the four verdicts hold under both answer orderings, with only comprehensiveness being position-sensitive, indicating a robust rather than order-driven preference. The advantage stems from SpectRAG's explicit evaluation and visualization discussion, which the LightRAG answer omits. \\
    \bottomrule
  \end{tabular*}

  \vspace{2pt}
  \raggedright\footnotesize{Pairwise verdicts from the \texttt{gpt-4o} judge (Answer 1 = SpectRAG, Answer 2 = LightRAG). Answers are condensed from the model outputs; full text is available in the supplementary material.}
\end{table*}
\subsection{RQ3}
SpectRAG matches the graph-RAG baseline and clearly outperforms the vector-RAG baseline on objective answer accuracy. On MultiHopRAG's 2{,}556 gold-answer questions, SpectRAG attains an $EM$ of 71.2 and $F1$ of 71.6, exceeding NaiveRAG ($EM$ 67.8 / $F1$ 68.3) by 3.4 $EM$ points (95\% CI [2.07, 4.73]), and is on par with LightRAG ($EM$ 70.5 / $F1$ 70.9), marginally ahead. Notably, SpectRAG's large retrieval advantage over LightRAG ($Recall@5$ 64.7 versus 49.2) translates into only a small answer-accuracy gain. This reflects the retrieval and generation gap: multi-hop answers are short (often yes/no or a single entity) and reachable from partial evidence, and all systems share the same generator model, temperature 0, identical prompt), so beyond a sufficiency threshold, additional retrieved evidence yields diminishing returns on final accuracy. SpectRAG's contribution is therefore a stronger, substantially cheaper retriever (RQ1), delivering answer accuracy that matches the state-of-the-art graph-RAG system while decisively surpassing standard vector retrieval. Detailed answer-accuracy results are reported in Section~\ref{sec:generation-results} (generation-level results).

\subsection{RQ4}
SpectRAG's spectral and PPR diffusion surfaces substantially more gold evidence, but the gain is concentrated in depth rather than in early rank. Against LightRAG, SpectRAG improves mean reciprocal rank ($MRR$) from 61.5 to 70.4 and recall at every cutoff ($Recall@2$ 39.1 versus 31.2, $Recall@5$ 64.7 versus 49.2, $Recall@10$ 84.0 versus 61.8). Against NaiveRAG, the result is a trade-off: NaiveRAG places its first gold document slightly earlier ($MRR$ 72.8 versus 70.4; $Recall@2$ 40.9 versus 39.1), whereas SpectRAG leads from $Recall@5$ onward (64.7 versus 62.4), with the gap widening at $Recall@10$ (84.0 versus 69.5). The same trade-off holds at the document level: SpectRAG's overall recall is 22.3 points higher than NaiveRAG's (91.8 versus 69.5), at the cost of 17.6 points of precision. Hybrid achieves the strongest early-rank results ($MRR$ 79.1, $Recall@2$ 46.3, $Recall@5$ 68.0), while SpectRAG attains the highest $Recall@10$, overall recall, and hit rate, retrieving at least one gold document for 99.9\% of questions versus 98.3\% for Hybrid and 96.1\% for NaiveRAG. SpectRAG therefore trades a small amount of early-rank precision for much deeper evidence coverage, which suits multi-hop questions whose answers depend on several pieces of evidence. Detailed retrieval-ranking results are reported in Section~\ref{sec:retrieval-results} (Retrieval-Level Results).

\section{Results and Analysis} \label{sec:results} %

\subsection{Main Results}
SpectRAG delivers two complementary advantages: an efficiency-at-parity result on MultiHopRAG and a quality result on UltraDomain. On MultiHopRAG, SpectRAG matches LightRAG's objective answer accuracy and outperforms NaiveRAG by 3.4 $EM$ points (\autoref{sec:generation-results}), while retrieving more gold evidence and covering it more deeply in the retrieved context (Section~{\ref{sec:retrieval-results}})  all at substantially lower index cost, evidenced by a markedly smaller graph and on-disk footprint. On UltraDomain, SpectRAG wins the pairwise LLM-judge evaluation against NaiveRAG, HyDE, and Hybrid, remains competitive with LightRAG, and shows a clear advantage in answer diversity (Tables~{\ref{tab:winrates-config1}} and~{\ref{tab:winrates-config2}}); a representative example is examined in Table~\ref{tab:cs_case_study}. These results answer the four research questions as follows. The four research questions are answered as follows. SpectRAG matches or exceeds LightRAG’s retrieval with a much smaller index (RQ1). Its holistic answer quality is close to LightRAG’s and ahead only on diversity in both configurations (RQ2). Its $EM$ and $F1$ are the highest of the five systems, 0.7 $EM$ points above LightRAG and 3.4 above NaiveRAG (RQ3). It retrieves more gold evidence than every baseline at depth 10, while Hybrid and NaiveRAG place the first gold document earlier (RQ4).
\label{sec:retrieval-results}

SpectRAG demonstrates a substantial improvement over LightRAG in retrieval quality, as evidenced by the ranking of gold evidence and recall across all evaluated cutoffs. Table~\ref{tab:retrieval-results} reports $MRR$ and $Recall@k$ on MultiHopRAG. SpectRAG attains an $MRR$ of 70.4, compared to 61.5 for LightRAG, and outperforms LightRAG across all recall cutoffs ($Recall@2$, $Recall@5$, and $Recall@10$: 39.1/64.7/84.0 versus 31.2/49.2/61.8). This advantage is not confined to aggregate performance but extends across all reasoning-intensive question types: $Recall@5$ favors SpectRAG on comparison (72.0 versus 54.8), inference (53.7 versus 38.8), and temporal (69.5 versus 55.5) questions. This consistency indicates that the observed retrieval gain reflects a general property of the retrieval mechanism rather than an artifact of any particular question type. Taken together with the ranking results, these findings address RQ4 and the retrieval component of RQ1, demonstrating that SpectRAG retrieves gold evidence more frequently and ranks it higher in context than LightRAG, consistently across question types.

Table~\ref{tab:retrieval-results} also reports Hybrid, HyDE and NaiveRAG. Hybrid has the best early-rank scores ($Recall@2$ 46.3, $Recall@5$ 68.0, $MRR$ 79.1), and NaiveRAG has a higher $MRR$ than SpectRAG (72.8 versus 70.4); SpectRAG has the highest overall recall (91.8) and $Recall@10$ (84.0).\footnote{NaiveRAG’s chunk-level retrieval maps only approximately onto the graph-node attribution used for LightRAG and SpectRAG, so its figures may be a lower bound.}

\begin{table}[!h]
\caption{Retrieval ranking quality on MultiHopRAG.}
\label{tab:retrieval-results}
\centering
\footnotesize
\renewcommand{\arraystretch}{1.3}
\setlength{\tabcolsep}{3.5pt}
\begin{tabular*}{\columnwidth}{@{\extracolsep{\fill}}lrrrrr@{}}
  \toprule
  \textbf{System} & \textbf{$Recall$} & \textbf{$R@2$} & \textbf{$R@5$} & \textbf{$R@10$} & \textbf{$MRR$} \\
  \midrule
  LightRAG & 65.6 & 31.2 & 49.2 & 61.8 & 61.5 \\
  Hybrid & 76.1 & \textbf{46.3} & \textbf{68.0} & 76.1 & \textbf{79.1} \\
  HyDE & 64.1 & 36.1 & 58.2 & 64.1 & 65.5 \\
  NaiveRAG & 69.5 & 40.9 & 62.4 & 69.5 & 72.8 \\ \hline
  \textbf{SpectRAG} & \textbf{91.8} & 39.1 & 64.7 & \textbf{84.0} & 70.4 \\
  \bottomrule
\end{tabular*}

\vspace{2pt}
\raggedright\footnotesize{Retrieval ranking quality on MultiHopRAG. $Recall$ is overall recall; $R@k$ is $Recall@k$ (fraction of gold evidence within the top-$k$ retrieved documents); $MRR$ is the reciprocal rank of the first gold-evidence document. Best result per column in bold.}
\end{table}
\subsection{Generation-Level Results}
\label{sec:generation-results}

This study aims to investigate end-to-end answer accuracy by employing a common generator across all five systems. This standardized approach ensures that any observed variations in performance are indicative of the retrieval quality, rather than discrepancies in the generative process itself. The overall accuracy reported on MultiHopRAG's 2{,}556 gold-answer questions demonstrates the evaluation's relevance and credibility. Objective accuracy is directly comparable across all baselines here, including NaiveRAG. Table~\ref {tab:overall-accuracy} reports overall accuracy on the MultiHopRAG dataset.

\begin{table}[!h]
\caption{Overall answer accuracy on MultiHopRAG dataset.}
\label{tab:overall-accuracy}
\centering
\footnotesize
\renewcommand{\arraystretch}{1.3}
\setlength{\tabcolsep}{6pt}
\begin{tabular*}{\columnwidth}{@{\extracolsep{\fill}}lrrr@{}}
  \toprule
  \textbf{System} & \textbf{$EM$} & \textbf{$F1$} & \textbf{$Contains$} \\
  \midrule
  HyDE & 65.2 & 65.6 & 65.4 \\
  NaiveRAG & 67.8 & 68.3 & 68.3 \\
  LightRAG & 70.5 & 70.9 & 70.8 \\
  Hybrid & 70.6 & 71.1 & 71.0 \\
  \hline
  \textbf{SpectRAG} & \textbf{71.2} & \textbf{71.6} & \textbf{71.6} \\
  \bottomrule
\end{tabular*}

\vspace{2pt}
\raggedright\footnotesize{$EM$ is normalized exact match; $F1$ is token overlap; $Contains$ indicates the gold answer appears as a whole-word span in the prediction. Best result per column in bold, ordered in ascending order according to the $EM$ values.}
\end{table}

SpectRAG attains the highest overall accuracy of all five systems on every metric ($EM$ 71.2, $F1$ 71.6, and $Contains$ 71.6). The margin over the next-best systems, LightRAG and Hybrid (both near 70.6), is narrow, while the gap over NaiveRAG (67.8) and HyDE (65.2) is substantial. Unlike the retrieval-level comparison, where SpectRAG's advantage was large, this is a clean but marginal win across the board. This is precisely the retrieval–generation gap in action: because all systems share the same generator and multi-hop answers are typically short (a yes/no verdict or a single entity) and often recoverable from partial evidence, retrieval improvements compress once a sufficiency threshold is reached, and additional or better-ordered evidence yields diminishing returns on final accuracy.

To make this compression concrete, Table~\ref{tab:per-type-accuracy} breaks accuracy down by question type. The per-type view shows that the small aggregate margin masks a mix of clear wins and near-ties.

\begin{table}[!h]
\caption{Per-type answer accuracy on MultiHopRAG.}
\label{tab:per-type-accuracy}
\centering
\footnotesize
\renewcommand{\arraystretch}{1.3}
\setlength{\tabcolsep}{4pt}
\begin{tabular*}{\columnwidth}{@{\extracolsep{\fill}}lrrrrr@{}}
  \toprule
  \textbf{Type} & \textbf{SpectRAG} & \textbf{LightRAG} & \textbf{Hybrid} & \textbf{HyDE} & \textbf{NaiveRAG} \\
  \midrule
  Comparison & 62.7 & \textbf{66.4} & 63.1 & 55.7 & 59.8 \\
  Inference & \textbf{95.1} & 92.2 & 92.2 & 87.0 & 87.7 \\
  Temporal & 43.9 & \textbf{44.9} & 42.7 & 35.8 & 40.1 \\
  Null & 83.1 & 72.8 & 87.4 & \textbf{89.7} & 89.7 \\
  \bottomrule
\end{tabular*}

\vspace{2pt}
\raggedright\footnotesize{Exact-match accuracy by question type. Best result per row in bold. Null questions are those with no answerable gold evidence; higher scores reflect correct abstention.}
\end{table}

The per-type breakdown clarifies where SpectRAG's aggregate win originates. SpectRAG dominates on inference questions (95.1, roughly three points above LightRAG and Hybrid and eight above the vector baselines), the type most dependent on assembling evidence across multiple hops, which is exactly where a stronger, better-ordered retriever should help. On temporal questions the three strongest systems are within 2.2 points of each other (42.7--44.9). Comparison questions separate the systems more: LightRAG leads with 66.4, 3.7 points above SpectRAG (62.7), and HyDE is lowest at 55.7. On these two types, SpectRAG's retrieval advantage over LightRAG does not carry over to answer accuracy. The null category behaves differently: it rewards correct abstention rather than evidence assembly, and here the vector baselines (89.7) and Hybrid (87.4) lead, while SpectRAG (83.1) still comfortably exceeds LightRAG (72.8). SpectRAG's lower null score reflects its tendency to surface candidate evidence even for unanswerable queries, a direct consequence of its broader retrieval, and represents the one regime where more aggressive retrieval is mildly counterproductive.

Considered collectively, the generation-level results confirm the paper's central claim: SpectRAG's value is a stronger and substantially cheaper retriever (RQ1), which converts into the best overall answer accuracy among all five systems and a decisive lead on the most retrieval-intensive question type, while the shared generator and short answer format cap how large that lead can become on the remaining types.

\subsection{Diversity, Redundancy, and Connectivity Analysis}
\label{sec:diversity-analysis}

To assess whether SpectRAG's denser, spectrally-organized chunk graph produces qualitatively richer answers, we evaluate SpectRAG (SpectRAG) against four baselines (NaiveRAG, HyDE, Hybrid, and LightRAG) using the UltraDomain-style pairwise win-rate protocol, in which a \texttt{gpt-4o} judge selects the preferred answer across four dimensions: comprehensiveness, diversity, empowerment, and overall quality. Each pair of columns reports the win rate (\%) of the named baseline versus SpectRAG on a given domain; the two columns in each pair sum to 100, and an SpectRAG value above 50 indicates SpectRAG is preferred. Table~\ref{tab:winrates-config1} reports the results under the first edge-signal configuration: semantic 0.300, section 0.250, entity 0.200, adjacency 0.084, utility question 0.083, and citation 0.083.

\begin{table*}[!h]
\caption{Win rates (\%) of baselines versus SpectRAG across four domains and four dimensions (signal configuration~1: semantic 0.300, section 0.250, entity 0.200, adjacency 0.084, utility question 0.083, citation 0.083)}
\label{tab:winrates-config1}
\centering
\footnotesize
\renewcommand{\arraystretch}{1.25}
\setlength{\tabcolsep}{4pt}
\begin{tabular*}{\textwidth}{@{\extracolsep{\fill}}lrrrrrrrrrr@{}}
  \toprule
  \textbf{Dimension} & \multicolumn{2}{c}{\textbf{Agriculture}} & \multicolumn{2}{c}{\textbf{CS}} & \multicolumn{2}{c}{\textbf{Legal}} & \multicolumn{2}{c}{\textbf{Mix}} & \multicolumn{2}{c}{\textbf{Average}} \\
  \cmidrule(lr){2-3}\cmidrule(lr){4-5}\cmidrule(lr){6-7}\cmidrule(lr){8-9}\cmidrule(lr){10-11}
  \multicolumn{11}{@{}l}{\textbf{Baseline: NaiveRAG}} \\
  \midrule
  & Base & \textbf{SpectRAG} & Base & \textbf{SpectRAG} & Base & \textbf{SpectRAG} & Base & \textbf{SpectRAG} & Base & \textbf{SpectRAG} \\
  Comprehensiveness & 43.8 & \textbf{56.2} & 37.5 & \textbf{62.5} & 51.3 & \textbf{48.7} & 37.5 & \textbf{62.5} & 42.5 & \textbf{57.5} \\
  Diversity & 32.5 & \textbf{67.5} & 21.3 & \textbf{78.7} & 52.5 & \textbf{47.5} & 37.5 & \textbf{62.5} & 35.9 & \textbf{64.1} \\
  Empowerment & 40.0 & \textbf{60.0} & 32.5 & \textbf{67.5} & 51.3 & \textbf{48.7} & 33.8 & \textbf{66.2} & 39.4 & \textbf{60.6} \\
  Overall & 40.0 & \textbf{60.0} & 35.0 & \textbf{65.0} & 51.3 & \textbf{48.7} & 35.0 & \textbf{65.0} & 40.3 & \textbf{59.7} \\
  \midrule
  \multicolumn{11}{@{}l}{\textbf{Baseline: HyDE}} \\
  \midrule
  & Base & \textbf{SpectRAG} & Base & \textbf{SpectRAG} & Base & \textbf{SpectRAG} & Base & \textbf{SpectRAG} & Base & \textbf{SpectRAG} \\
  Comprehensiveness & 37.5 & \textbf{62.5} & 36.8 & \textbf{63.2} & 48.8 & \textbf{51.2} & 30.0 & \textbf{70.0} & 38.3 & \textbf{61.7} \\
  Diversity & 22.5 & \textbf{77.5} & 25.0 & \textbf{75.0} & 43.8 & \textbf{56.2} & 22.5 & \textbf{77.5} & 28.5 & \textbf{71.5} \\
  Empowerment & 35.0 & \textbf{65.0} & 34.2 & \textbf{65.8} & 47.5 & \textbf{52.5} & 28.7 & \textbf{71.3} & 36.4 & \textbf{63.6} \\
  Overall & 35.0 & \textbf{65.0} & 34.2 & \textbf{65.8} & 47.5 & \textbf{52.5} & 28.7 & \textbf{71.3} & 36.4 & \textbf{63.6} \\
  \midrule
  \multicolumn{11}{@{}l}{\textbf{Baseline: Hybrid}} \\
  \midrule
  & Base & \textbf{SpectRAG} & Base & \textbf{SpectRAG} & Base & \textbf{SpectRAG} & Base & \textbf{SpectRAG} & Base & \textbf{SpectRAG} \\
  Comprehensiveness & 47.5 & \textbf{52.5} & 41.2 & \textbf{58.8} & 50.0 & \textbf{50.0} & 42.5 & \textbf{57.5} & 45.3 & \textbf{54.7} \\
  Diversity & 41.2 & \textbf{58.8} & 28.7 & \textbf{71.3}& 46.3 & \textbf{53.7} & 41.2 & \textbf{58.8} & 39.4 & \textbf{60.6} \\
  Empowerment & 45.0 & \textbf{55.0} & 38.7 & \textbf{61.3} & 48.8 & \textbf{51.2} & 37.5 & \textbf{62.5} & 42.5 & \textbf{57.5} \\
  Overall & 46.3 & \textbf{53.7} & 38.7 & \textbf{61.3} & 48.8 & \textbf{51.2} & 41.2 & \textbf{58.8} & 43.8 & \textbf{56.2} \\
  \midrule
  \multicolumn{11}{@{}l}{\textbf{Baseline: LightRAG}} \\
  \midrule
  & Base & \textbf{SpectRAG} & Base & \textbf{SpectRAG} & Base & \textbf{SpectRAG} & Base & \textbf{SpectRAG} & Base & \textbf{SpectRAG} \\
  Comprehensiveness & 43.8 & \textbf{56.2} & 56.2 & \textbf{43.8} & 47.5 & \textbf{52.5} & 53.7 & \textbf{46.3} & 50.3 & \textbf{49.7} \\
  Diversity & 35.0 & \textbf{65.0} & 52.5 & \textbf{47.5} & 45.0 & \textbf{55.0} & 47.5 & \textbf{52.5} & 45.0 & \textbf{55.0} \\
  Empowerment & 42.5 & \textbf{57.5} & 58.8 & \textbf{41.2} & 48.8 & \textbf{51.2} & 56.2 & \textbf{43.8} & 51.6 & \textbf{48.4} \\
  Overall & 42.5 & \textbf{57.5} & 58.8 & \textbf{41.2} & 48.8 & \textbf{51.2} & 55.0 & \textbf{45.0} & 51.3 & \textbf{48.7} \\
  \bottomrule
\end{tabular*}

\vspace{2pt}
\raggedright\footnotesize{Each pair of columns gives the win rate (\%) of the baseline (Base) versus SpectRAG; the two sum to 100, and SpectRAG values above 50 (bold in the Average column) indicate SpectRAG is preferred. Judged by \texttt{gpt-4o}. Signal weights sum to 1.000 and are fixed across all domains and baselines.}
\end{table*}

Two patterns are consistent under this configuration. First, \textbf{diversity is SpectRAG's strongest dimension against every baseline}: it wins on average 64.1 against NaiveRAG, 71.5 against HyDE, 60.6 against Hybrid, and 55.0 against LightRAG. This is the signature of a retriever that spreads its evidence across structurally distinct regions of the graph rather than concentrating it in a small, redundant neighborhood. Second, \textbf{the margin scales with how redundancy-prone the baseline is}: SpectRAG wins all four dimensions decisively against HyDE and NaiveRAG, leads throughout against Hybrid, and against LightRAG is close to even on comprehensiveness, empowerment, and overall (49.7, 48.4, 48.7) while still winning diversity (55.0). Even when the two graph-based systems are matched on aggregate quality, SpectRAG's denser chunk graph continues to deliver more varied, less redundant evidence, consistent with its higher edges-per-node ratio (RQ1).

To test whether this behavior depends on the exact signal weighting, we repeat the evaluation under a second, contrasting configuration that shifts weight away from raw semantic similarity and toward utility-question matching and entity structure: Semantic 0.150, Section 0.200, entity 0.250, adjacency 0.100, utility question 0.280, and citation 0.020. The results are reported in Table~\ref{tab:winrates-config2}.

\begin{table*}[t]
\caption{Win rates (\%) of baselines versus SpectRAG across four domains and four dimensions (signal configuration~2: semantic 0.150, section 0.200, entity 0.250, adjacency 0.100, utility question 0.280, citation 0.020)}
\label{tab:winrates-config2}
\centering
\footnotesize
\renewcommand{\arraystretch}{1.25}
\setlength{\tabcolsep}{4pt}
\begin{tabular*}{\textwidth}{@{\extracolsep{\fill}}lrrrrrrrrrr@{}}
  \toprule
  \textbf{Dimension} & \multicolumn{2}{c}{\textbf{Agriculture}} & \multicolumn{2}{c}{\textbf{CS}} & \multicolumn{2}{c}{\textbf{Legal}} & \multicolumn{2}{c}{\textbf{Mix}} & \multicolumn{2}{c}{\textbf{Average}} \\
  \cmidrule(lr){2-3}\cmidrule(lr){4-5}\cmidrule(lr){6-7}\cmidrule(lr){8-9}\cmidrule(lr){10-11}
  \multicolumn{11}{@{}l}{\textbf{Baseline: NaiveRAG}} \\
  \midrule
  & Base & \textbf{SpectRAG} & Base & \textbf{SpectRAG} & Base & \textbf{SpectRAG} & Base & \textbf{SpectRAG} & Base & \textbf{SpectRAG} \\
  Comprehensiveness & 43.8 & \textbf{56.2} & 41.2 & \textbf{58.8} & 53.7 & \textbf{46.3} & 43.8 & \textbf{56.2} & 45.6 & \textbf{54.4} \\
  Diversity & 37.5 & \textbf{62.5} & 26.2 & \textbf{73.8} & 55.0 & \textbf{45.0} & 41.2 & \textbf{58.8} & 40.0 & \textbf{60.0} \\
  Empowerment & 40.0 & \textbf{60.0} & 36.3 & \textbf{63.7} & 51.3 & \textbf{48.7} & 41.2 & \textbf{58.8} & 42.2 & \textbf{57.8} \\
  Overall & 41.2 & \textbf{58.8} & 36.3 & \textbf{63.7} & 52.5 & \textbf{47.5} & 42.5 & \textbf{57.5} & 43.1 & \textbf{56.9} \\
  \midrule
  \multicolumn{11}{@{}l}{\textbf{Baseline: HyDE}} \\
  \midrule
  & Base & \textbf{SpectRAG} & Base & \textbf{SpectRAG} & Base & \textbf{SpectRAG} & Base & \textbf{SpectRAG} & Base & \textbf{SpectRAG} \\
  Comprehensiveness & 40.0 & \textbf{60.0} & 32.5 & \textbf{67.5} & 48.8 & \textbf{51.2} & 25.0 & \textbf{75.0} & 36.6 & \textbf{63.4} \\
  Diversity & 27.5 & \textbf{72.5} & 22.5 & \textbf{77.5} & 41.2 & \textbf{58.8} & 18.8 & \textbf{81.2} & 27.5 & \textbf{72.5} \\
  Empowerment & 33.8 & \textbf{66.2} & 30.0 & \textbf{70.0} & 48.8 & \textbf{51.2} & 22.5 & \textbf{77.5} & 33.8 & \textbf{66.2} \\
  Overall & 35.0 & \textbf{65.0} & 30.0 & \textbf{70.0} & 48.8 & \textbf{51.2} & 22.5 & \textbf{77.5} & 34.1 & \textbf{65.9} \\
  \midrule
  \multicolumn{11}{@{}l}{\textbf{Baseline: Hybrid}} \\
  \midrule
  & Base & \textbf{SpectRAG} & Base & \textbf{SpectRAG} & Base & \textbf{SpectRAG} & Base & \textbf{SpectRAG} & Base & \textbf{SpectRAG} \\
  Comprehensiveness & 46.3 & \textbf{53.7} & 51.3 & \textbf{48.7} & 47.5 & \textbf{52.5} & 41.2 & \textbf{58.8} & 46.6 & \textbf{53.4} \\
  Diversity & 46.3 & \textbf{53.7} & 47.5 & \textbf{52.5} & 36.3 & \textbf{63.7} & 35.0 & \textbf{65.0} & 41.2 & \textbf{58.8} \\
  Empowerment & 47.5 & \textbf{52.5} & 50.0 & \textbf{50.0} & 45.0 & \textbf{55.0} & 41.2 & \textbf{58.8} & 45.9 & \textbf{54.1} \\
  Overall & 47.5 & \textbf{52.5} & 50.0 & \textbf{50.0} & 45.0 & \textbf{55.0} & 40.0 & \textbf{60.0} & 45.6 & \textbf{54.4} \\
  \midrule
  \multicolumn{11}{@{}l}{\textbf{Baseline: LightRAG}} \\
  \midrule
  & Base & \textbf{SpectRAG} & Base & \textbf{SpectRAG} & Base & \textbf{SpectRAG} & Base & \textbf{SpectRAG} & Base & \textbf{SpectRAG} \\
  Comprehensiveness & 45.0 & \textbf{55.0} & 57.5 & \textbf{42.5} & 46.3 & \textbf{53.7} & 53.7 & \textbf{46.3} & 50.6 & \textbf{49.4} \\
  Diversity & 33.8 & \textbf{66.2} & 58.8 & \textbf{41.2} & 37.5 & \textbf{62.5} & 46.3 & \textbf{53.7} & 44.1 & \textbf{55.9} \\
  Empowerment & 43.8 & \textbf{56.2} & 55.0 & \textbf{45.0} & 46.3 & \textbf{53.7} & 50.0 & \textbf{50.0} & 48.8 & \textbf{51.2} \\
  Overall & 43.8 & \textbf{56.2} & 55.0 & \textbf{45.0} & 46.3 & \textbf{53.7} & 50.0 & \textbf{50.0} & 48.8 & \textbf{51.2} \\
  \bottomrule
\end{tabular*}

\vspace{2pt}
\raggedright\footnotesize{Each pair of columns gives the win rate (\%) of the baseline (Base) versus SpectRAG; the two sum to 100, and SpectRAG values above 50 (bold in the Average column) indicate SpectRAG is preferred. Judged by \texttt{gpt-4o}. Signal weights sum to 1.000 and are fixed across all domains and baselines.}
\end{table*}

Comparing the two configurations yields two findings. First, \textbf{the diversity advantage is stable across both weightings}: diversity remains SpectRAG's strongest dimension against every baseline (60.0, 72.5, 58.8, 55.9), with margins that closely track configuration~1. Because the two weightings emphasize almost disjoint signal sets yet produce the same diversity ordering, the effect is a structural property of the denser chunk graph rather than an artifact of any single signal. Second, \textbf{the utility-question-weighted configuration strengthens SpectRAG against the hardest baseline}: against LightRAG, SpectRAG moves from winning only diversity (configuration~1) to additionally winning empowerment (51.2) and the overall verdict (51.2), losing only comprehensiveness by a narrow margin (49.4). The gains against the weaker baselines are preserved throughout. This indicates that anchoring retrieval on utility questions and entity structure, rather than surface semantic similarity, helps most precisely where the competition is strongest. Together, the two configurations bound SpectRAG's qualitative behavior: its diversity and redundancy advantages are insensitive to the exact signal weighting. At the same time, its aggregate standing against LightRAG can be tuned from roughly even to a slight lead.

\section{Discussion}

The experimental findings provide several observations about the behavior  provide several observations about SpectRAG's yield provide several observations about the behavior of SpectRAG in multi-hop and graph-structured retrieval settings. Overall, SpectRAG shows its strongest advantage when the task requires broad evidence recovery rather than only early precision in a ranked list. In the MultiHopRAG retrieval-ranking evaluation, SpectRAG achieves the highest overall recall and the highest $Recall@10$, indicating that the proposed graph expansion and spectral diffusion stages help recover a wider set of gold evidence. However, Hybrid retrieval obtains the best $Recall@2$, $Recall@5$, and $MRR$, suggesting that SpectRAG does not always place the first relevant evidence item earlier than strong lexical--dense baselines. This pattern is consistent with the design of SpectRAG: the method is not optimized solely for nearest-neighbor ranking, but for constructing a broader, more structurally useful candidate pool before final context selection.

The generation-level results further indicate that improvements in retrieval coverage can translate into modest but consistent gains in answer accuracy. On MultiHopRAG, SpectRAG obtains the highest $EM$, $F1$, and answer-containment scores among the compared systems. The margin is not large, but it is consistent across all three generation metrics. This recommends that the final context selected by SpectRAG can provide useful evidence to the generator, especially when the answer depends on information distributed across multiple passages. Simultaneously, the relatively small generation margin highlights that better retrieval does not automatically guarantee substantially better generation. The LLM must correctly integrate the supplied evidence, follow the reasoning path, and avoid relying on irrelevant or weakly connected context.

Based on the per-type analysis, it is evident that SpectRAG is especially useful in inference-type questions, scoring the highest $EM$ scores amongst all the evaluated systems. These results confirm our hypothesis that graph-spectral representations are beneficial when the question requires connecting evidence pieces rather than finding a direct match. The results for the comparison and temporal questions are mixed, with LightRAG slightly better than SpectRAG. This implies that some question types could be helped more by other relational representations or different kinds of prompts. For null questions, SpectRAG abstains less reliably than the dense baselines (83.1 versus 89.7 for NaiveRAG and HyDE, and 87.4 for Hybrid), although it still exceeds LightRAG (72.8); as discussed in Section~{\ref{sec:generation-results}}, graph expansion can surface related but insufficient context, which discourages abstention. Beyond retrieval quality, SpectRAG also incurs a lower memory cost than LightRAG. This advantage comes from node count and index size rather than edge count: on MultiHopRAG, SpectRAG builds 5,022 nodes against LightRAG's 15,803 and persists a 46~MB index against 274~MB, although its chunk graph is denser per node (3.85 versus 1.07 edges per node; Table~4). The observed gains are therefore achieved with a substantially smaller index.

The pairwise LLM-judge evaluations across agriculture, CS, legal, and mix domains provide a complementary view of generation quality. Across two signal configurations, SpectRAG generally achieves favorable win rates against NaiveRAG, HyDE, and Hybrid baselines, especially in diversity and empowerment. The diversity gains are the most stable pattern: SpectRAG frequently receives higher diversity scores across domains and configurations. This is aligned with the method's objective of reducing redundant evidence and selecting context from multiple graph-spectral regions. The results against LightRAG are more nuanced. SpectRAG is competitive and often stronger in diversity, but it does not consistently dominate LightRAG across all domains, particularly in the CS domain. This indicates that graph-spectral selection improves some aspects of answer quality, but its advantage depends on the corpus structure, baseline strength, and signal weighting.

The qualitative case study provides an interpretable example of this behavior. For the anomaly detection question, SpectRAG is preferred in comprehensiveness, diversity, empowerment, and overall quality. The judged explanation suggests that SpectRAG covers a wider set of methods, domains, evaluation strategies, and visualization tools, while LightRAG provides a more compact but less varied synthesis. Although this is only a single example, it illustrates the kind of context composition that SpectRAG is intended to support: a final answer grounded in multiple complementary evidence aspects rather than a narrow set of locally similar passages.

\subsection{Why Spectral Structure Helps}%refer to other types of rag systems
The results suggest that spectral graph structure is particularly useful when retrieval requires broad evidence coverage and relational coherence. Existing RAG pipelines exploit corpus structure only partially: purely embedding-based methods select chunks that are locally similar to the query without modeling the relationships among candidate chunks, whereas graph-based methods encode relations at the entity level and traverse them through local neighborhood expansion. In neither case is the collection of retrieval units treated as a global structure whose connectivity governs selection. Through a weighted chunk graph, SpectRAG instead incorporates semantic similarity, document adjacency, section hierarchy, entity co-occurrence, citation relations, and utility-question similarity as complementary signals of corpus structure.

The Laplacian spectral coordinates offer a compact description of such graph topology. Low-frequency spectral coordinates describe global connectivity patterns and communities in the graph, allowing detection of chunks belonging to the same group despite differences in their direct similarity. The use of the spectral coordinates in SpectRAG is crucial for the cluster-aware selection because it allows to avoid domination of one local neighborhood in the resulting context set.

Spectral diffusion also contributes to the retrieval process by propagating query relevance over the induced candidate subgraph. A chunk that is not highly similar to the query may still receive increased relevance if it is strongly connected to directly retrieved evidence. This mechanism is particularly important for bridge evidence, where an intermediate passage connects two otherwise disconnected pieces of information. The high overall recall and $Recall@10$ observed on MultiHopRAG are consistent with this behavior: SpectRAG appears to recover a broader evidence set, even when some of that evidence is not ranked at the very top.

\subsection{Practical Deployment in RAG Pipeline and Framework}%refer instead of explain

\begin{figure*}
\centering
\includegraphics[width=0.9\textwidth]{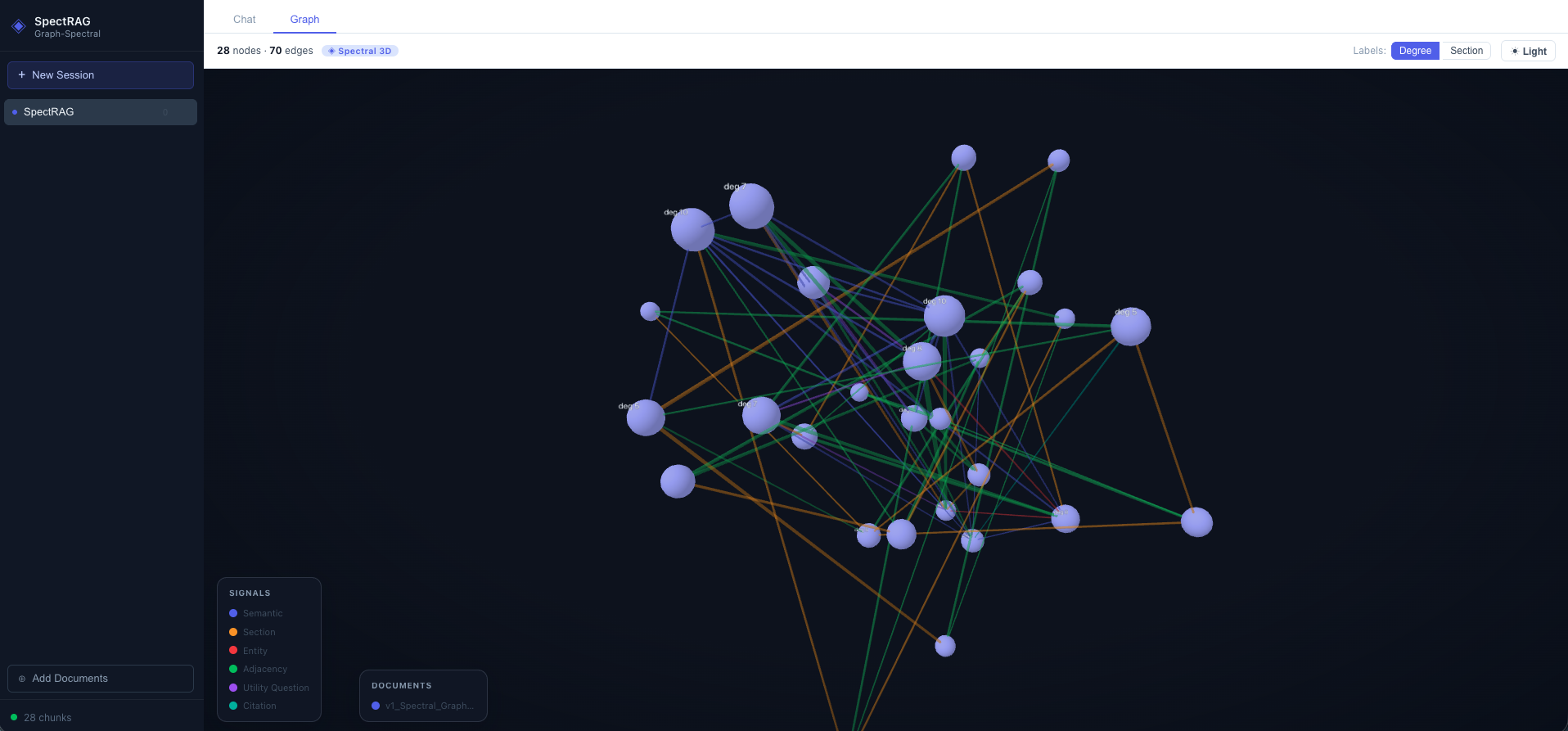}
\caption{Example graph-view interface of the proposed SpectRAG framework. Nodes represent document chunks and edges represent weighted relation signals, including semantic similarity, section relations, entity overlap, adjacency, utility-question similarity, and citation links.}
\label{fig:graph-framework}
\end{figure*}

In the current implementation, dense candidate seeding and utility-question retrieval are performed using exact flat inner-product search with FAISS \texttt{IndexFlatIP}. This choice avoids approximation error and makes the preliminary evaluation easier to interpret, since differences between systems are not caused by approximate nearest-neighbor indexing. For small and medium-sized experimental corpora, exact search is also practical and reproducible. In larger deployments, however, this component may become a bottleneck and can be replaced by approximate indexing methods without changing the downstream graph expansion, diffusion, and token-budget limit stages.

Most graph and spectral computations can be moved to the offline indexing stage. Chunk embeddings, graph construction, Laplacian computation, and spectral coordinates can be precomputed and cached. At query time, SpectRAG only needs to retrieve a candidate seed set, expand it through graph neighborhoods, construct the induced candidate subgraph, diffuse relevance within that subgraph, and select the final context. This separation between offline indexing and online retrieval makes the method practical for RAG systems where the corpus changes less frequently than the query stream.

In addition to the retrieval pipeline, SpectRAG includes a graph-view interface for inspecting the constructed chunk graph and its spectral organization. Fig.~\ref{fig:graph-framework} shows an example visualization of the weighted chunk graph generated from an indexed document collection. In this view, nodes correspond to document chunks and edges correspond to relation signals used during graph construction, including semantic similarity, section-level relations, entity overlap, document adjacency, utility-question similarity, and citation links. The node layout provides an interpretable view of how chunks are organized in the graph, while edge colors indicate the dominant relation types connecting different parts of the corpus.

Each node in the graph is associated with degree information that summarizes its local connectivity. In case of the chunk node $c_i$, the unweighted degree refers to the number of other chunks that connect with the current one, while the weighted degree is the sum of weights of the edges. Since SpectRAG builds a weighted chunk graph, the weighted degree is:

\begin{equation}
\deg(c_i)=\sum_{j=1}^{n} a_{ij},
\end{equation}

Using the graph visualization, it is possible to identify dense local neighborhoods, bridge nodes, isolated regions, and inter-document connections. These aspects are critical in establishing that the graph that is constructed properly captures the structure of the corpus before graph expansion, spectral relevance propagation, and budgeting of tokens for context selection are done. In actual RAG implementations, this visualization can help in identifying retrieval problems and graph sparsity.

\section{Limitations and Threats to Validity} \label{limitations}

Several limitations should be considered when interpreting the results of this study.

\textbf{Retrieval--generation gap.}
Although SpectRAG improves retrieval coverage in several settings, retrieval gains do not always lead to proportional improvements in generated answers. The generator may fail to use retrieved evidence correctly, may ignore bridge passages, or may be sensitive to the ordering of the final context. This is especially important because the answer accuracy results show only modest improvements over strong baselines. Therefore, retrieval-level and generation-level metrics should be interpreted separately.

\textbf{Task-dependent performance.}
SpectRAG does not dominate all question types. The per-type evaluation shows strong performance on inference questions, but weaker or mixed results on comparison, temporal, and null questions. This indicates that graph-spectral retrieval is most beneficial when the task requires evidence connection and synthesis, but may be less effective when the task depends on precise temporal ordering, direct comparison, or abstention calibration.

\textbf{Signal-weight sensitivity.}
The two evaluated signal configurations indicate that performance can change depending on the relative weights assigned to semantic similarity, section hierarchy, entity overlap, adjacency, utility-question similarity, and citation links. While SpectRAG remains generally competitive across configurations, some domains and baselines respond differently to weight changes. This creates a threat to external validity because a configuration that works well for one corpus may not be optimal for another.

\textbf{Chunking sensitivity.}
The method is evaluated under token-based chunking configurations, including shorter and longer chunk granularities. Chunk size and overlap affect graph topology, candidate retrieval, redundancy, and the amount of local context available in each selected unit. Larger chunks preserve more local context but may reduce the benefit of graph-based composition. Therefore, chunking should be treated as an experimental variable rather than a fixed preprocessing choice.

\textbf{Computational scalability.}
The current implementation uses exact brute-force inner-product search through FAISS \texttt{IndexFlatIP} for both chunk embedding retrieval and utility-question retrieval. This avoids approximation error but has higher query-time cost than approximate nearest-neighbor indexing. In addition, spectral computation over large graphs can be expensive. Practical large-scale deployments may require sparse graph construction, approximate eigensolvers, graph coarsening, caching, or approximate vector indexes.

\section{Conclusion}
This paper introduced SpectRAG, a spectral graph-augmented retrieval framework for constructing diverse, connected, and evidence-complete context sets in retrieval-augmented generation. The central motivation of the work is that standard dense retrieval is effective for identifying passages that are individually similar to a query, but it does not explicitly optimize the collective utility of the retrieved context. This limitation becomes more important in multi-hop and cross-document settings, where the answer may depend on evidence distributed across several passages, documents, or structural relations.

SpectRAG addresses this problem by combining dense candidate seeding, weighted chunk graph construction, graph neighborhood expansion, spectral relevance propagation, and token-budgeted final context selection. Rather than replacing dense retrieval, the proposed framework uses it as a high-recall entry point and then applies graph-spectral signals to recover supporting evidence, reduce redundant context, and improve structural coherence. The final context is assembled under a token budget, allowing the system to adapt the number of selected chunks according to evidence length and available context-window capacity.

The experimental results show that SpectRAG is most beneficial when retrieval requires broad evidence recovery and relational reasoning. On MultiHopRAG, SpectRAG achieves the highest overall recall and $Recall@10$, indicating that graph expansion and spectral propagation help recover a wider set of relevant evidence. In generation-level evaluation, SpectRAG obtains the strongest $EM$, $F1$, and answer-containment scores among the compared systems, although the margins remain moderate. These results suggest that improved evidence coverage can support better generation, but also confirm that retrieval gains do not automatically translate into large answer-quality gains unless the generator can effectively use the selected evidence.

SpectRAG also turns out to be an especially strong solution for inference problems and often helps diversify judgments in LLM-based pairwise evaluations of diversity. In the vast majority of domains and signal combinations, SpectRAG outperforms the NaiveRAG, HyDE, and Hybrid baselines across diversity, empowerment, and total performance metrics. Comparing the results of SpectRAG with those of LightRAG reveals some interesting conclusions: graph-spectral context selection performs better than the alternative in some cases but not in others. Against LightRAG, the results are inconclusive. The SpectRAG approach performs better in terms of information diversity under both signal configurations; however, its winning rate is 48.7\% under configuration 1 and 51.2\% under configuration 2. On the other hand, the LightRAG approach is the preferred choice in the CS realm under both signal configurations. Since the proposed approach ought not to be construed as a wholesale improvement over prevailing retrieval strategies, but rather as a complementary method, warranted specifically in settings where evidence diversity and connectivity must be preserved.

Overall, the findings support the main hypothesis of this work: graph-spectral structure can provide useful retrieval signals beyond local vector similarity. By representing chunks as nodes in a weighted graph and leveraging spectral structure during candidate expansion, relevance propagation, and token-budgeted context assembly, SpectRAG offers a principled approach to moving from independent passage ranking to graph-aware context construction. Future work should extend the evaluation to larger corpora and additional domains, conduct human preference studies, develop stronger graph-enhanced baselines, and explore scalable indexing variants. Further investigation is also needed to optimize signal weights, improve temporal and comparative reasoning, and examine how context ordering affects LLMs' ability to exploit graph-selected evidence.

\section*{Acknowledgments}
The authors received no financial support for the research, authorship, and/or publication of this article.

\section*{Code Availability}
The source code, configuration files, evaluation scripts, and documentation for SpectRAG will be made publicly available in a GitHub repository. To support reproducibility, the released code will include experiment configurations, dependency information, and instructions for reproducing the reported results. The repository is available at: \url{https://github.com/SightAndCite/SpectRAG}.

\end{document}